\documentclass[twoside,11pt]{article}
\usepackage{jmlr2e-mod}

\usepackage{amssymb,amsfonts,amsmath}
\usepackage{braket}
\usepackage{xparse}
\usepackage{graphicx} 
\usepackage[caption=false,font=footnotesize,justification=centering]{subfig} 
\usepackage{xcolor}
\usepackage{xfrac}
\usepackage{enumitem}
\usepackage{bm}
\setlist{noitemsep}
\usepackage{fontawesome}
\usepackage{fancyvrb}
\usepackage{tikz}
\usetikzlibrary{patterns}
\usetikzlibrary{decorations.markings,positioning,plotmarks,shapes.geometric}
\usepackage{pgfplots, pgfplotstable}
\usepgfplotslibrary{statistics} 
\pgfplotsset{compat=1.14}

\pgfplotsset{compat=1.11,
    /pgfplots/ybar legend/.style={
    /pgfplots/legend image code/.code={%
       \draw[##1,/tikz/.cd,yshift=-0.25em]
        (0cm,0cm) rectangle (3pt,0.8em);},
   },
}

\usepackage{hyperref}
\hypersetup{
  hypertexnames=false,
  colorlinks=true,
  allcolors=green!50!black,
  urlcolor=red!50!black
}

\usepackage{algorithm}
\usepackage{algpseudocode}

\usepackage[capitalize,nameinlink]{cleveref}

\makeatletter
\newcounter{algorithmicH}%
\let\oldalgorithmic\algorithmic
\renewcommand{\algorithmic}{%
  \stepcounter{algorithmicH}%
  \oldalgorithmic}%
\renewcommand{\theHALG@line}{ALG@line.\thealgorithmicH.\arabic{ALG@line}}
\makeatother

\makeatletter
\newcommand*{\transpose}{%
  {\mathpalette\@transpose{}}%
}
\newcommand*{\@transpose}[2]{%
  \raisebox{\depth}{$\m@th#1\intercal$}%
}
\makeatother

\NewDocumentCommand{\MatrixAlt}{m G{\BooleanFalse} G{\BooleanFalse}}{%
  \mathbf{ \IfBooleanTF{#2}{\hat}{} #1 }%
  \IfBooleanTF{#3}{^{\transpose}}{}%
}

\NewDocumentCommand{\Matrix}{m G{\BooleanFalse} G{\BooleanFalse} G{\BooleanFalse}}{%
  \IfBooleanTF{#4}%
  {\RV{ \MatrixAlt{#1}{#2}{#3} }}
  {     \MatrixAlt{#1}{#2}{#3}  }
}

\NewDocumentCommand{\ElementAlt}{m G{\BooleanFalse}}{%
  \IfBooleanTF{#2}{\hat}{} #1_{ij}%
}

\NewDocumentCommand{\Element}{m G{\BooleanFalse} G{\BooleanFalse}}{%
  \IfBooleanTF{#3}{ \RV{ \ElementAlt{#1}{#2} } }{ \ElementAlt{#1}{#2} }%
}

\NewDocumentCommand{\Row}{m O{} m m}{%
  \IfBooleanTF{#4}%
  {\RV{\bm{\mathbf{#1}}_{#2}\IfBooleanTF{#3}{^{\transpose}}{}}}%
  {   {\bm{\mathbf{#1}}_{#2}\IfBooleanTF{#3}{^{\transpose}}{}}}%
}

\NewDocumentCommand { \Good }{ } {\textcolor{green!50!black}{\faCheck}}
\NewDocumentCommand { \Bad }{ } {\textcolor{red!50!black}{\faTimes}}
\NewDocumentCommand { \MethodList } { s s } {
  \footnotesize
  \begin{tabular}{c@{\,}l}
    \IfBooleanTF{#1}{\Bad}{\Good} & PCA\\ 
    \IfBooleanTF{#2}{\Bad}{\Good} & COCA\\ 
    \Good & XPCA
  \end{tabular}
}

\newcommand{\doi}[1]{\href{http://dx.doi.org/#1}{#1}}

\definecolor{brightblue}{rgb}{0.1, 0.3, 1.0}
\NewDocumentCommand { \RV }{ m }{\textcolor{brightblue}{#1}}

\newcommand{\I}{\mathbf{I}}
\NewDocumentCommand{\D}{t' s}{ \Matrix{D}{\BooleanFalse}{#1}{#2} }
\DeclareDocumentCommand{\U}{t~ t' s}{ \Matrix{U}{#1}{#2}{#3} }
\NewDocumentCommand{\V}{t~ t' s}{ \Matrix{V}{#1}{#2}{#3} }
\NewDocumentCommand{\W}{t' s}{ \Matrix{W}{\BooleanFalse}{#1}{#2} }
\NewDocumentCommand{\X}{t~ t' s}{ \Matrix{X}{#1}{#2}{#3} }
\NewDocumentCommand{\T}{t~ t' s}{ \Matrix{\Theta}{#1}{#2}{#3} }
\NewDocumentCommand{\Y}{t' s}{ \Matrix{Y}{\BooleanFalse}{#1}{#2} }
\NewDocumentCommand{\Z}{t' s}{ \Matrix{Z}{\BooleanFalse}{#1}{#2} }

\NewDocumentCommand { \rowe } { t' s } { \Row{\epsilon}{#1}{#2} }
\NewDocumentCommand { \rowt } { t' s } { \Row{t}{#1}{#2} }
\NewDocumentCommand { \rowu } { t' s } { \Row{u}{#1}{#2} }
\NewDocumentCommand { \rowx } { t' s } { \Row{x}{#1}{#2} }
\NewDocumentCommand { \rowxi } { t' s } { \Row{x}{#1}{#2}_i }
\NewDocumentCommand { \roww } { t' s } { \Row{w}{#1}{#2} }
\NewDocumentCommand { \rowy } { t' s } { \Row{y}{#1}{#2} }
\NewDocumentCommand { \rowz } { t' s } { \Row{z}{#1}{#2} }

\NewDocumentCommand{\lij}{s}{\IfBooleanTF{#1}{\tilde}{}\ell_{ij}}
\NewDocumentCommand{\rij}{s}{\IfBooleanTF{#1}{\tilde}{}r_{ij}}

\NewDocumentCommand{\tij}{t~ s}{\Element{\theta}{#1}{#2}}
\NewDocumentCommand{\xij}{t~ s}{\Element{x}{#1}{#2}}
\NewDocumentCommand{\yij}{t~ s}{\Element{y}{#1}{#2}}
\NewDocumentCommand{\zij}{s}{\IfBooleanTF{#1}{\RV{z_{ij}}}{z_{ij}}}

\newcommand{\qtext}[1]{\quad\text{#1}\quad}

\newcommand{\Real}{\mathbb{R}}
\newcommand{\PosReal}{\Real_{+}}

\newcommand{\Fhat}{\ensuremath\hat F}
\newcommand{\FhatInv}{\ensuremath\hat F^{-1}}
\newcommand{\Falt}{\ensuremath\tilde F}
\newcommand{\FaltInv}{\ensuremath\tilde F^{-1}}
\newcommand{\PhiInv}{\Phi^{-1}}

\newcommand{\Cj}{\mathcal{C}_j}

\NewDocumentCommand { \Normal } { G{\mu} G{\sigma^2} } {\mathcal{N}(#1,#2)}
\NewDocumentCommand { \MultiNormal }{ O{} G{\bm{\mu}} G{\bm{\Sigma}} }{\mathcal{N}_{#1}(#2,#3)}

\NewDocumentCommand{\diffl}{s}{\bar{\ell}\IfBooleanTF{#1}{_{ij}}{}}
\NewDocumentCommand{\diffr}{s}{\bar{r}\IfBooleanTF{#1}{_{ij}}{}}
\NewDocumentCommand{\dpdf}{s}{\psi\IfBooleanTF{#1}{_{ij}}{}}
\NewDocumentCommand{\prob}{s}{P\IfBooleanTF{#1}{_{ij}}{}}
\NewDocumentCommand{\loss}{s}{L\IfBooleanTF{#1}{_{ij}}{}}
\NewDocumentCommand{\FD}{s m m}{%
  \frac{\IfBooleanTF{#1}{d}{\partial}{#2}}{\IfBooleanTF{#1}{d}{\partial}{#3}}}
\NewDocumentCommand{\SD}{s m m}{%
  \frac{\IfBooleanTF{#1}{d}{\partial}^2{#2}}{\IfBooleanTF{#1}{d}{\partial}{#3}^2}}
\NewDocumentCommand{\LA}{}{\mathbf{A}}
\NewDocumentCommand{\LB}{}{\mathbf{D}_i}
\NewDocumentCommand{\LC}{}{\mathbf{D}_j}

\DeclareMathOperator{\negloglik}{NLL}

\pgfplotsset{
  every axis/.append style={
    width = 2in, 
    height = 2in,
    clip = false,
    xlabel near ticks,
    ylabel near ticks,
    label style={font=\footnotesize},
    tick label style={font=\footnotesize},
  },
  exdist/.append style={
    width = 1.8in, 
    height = 1.5in,
    ticks=none,
    every axis plot/.append style={blue,very thick,no marks}
  },
  empdist/.append style={
    ymin=0,ymax=0.4,
    xmin=0,xmax=5.75,
    xtick = \edfx,
    xticklabels = {$v_1$, $v_2$, $v_3$, $v_4$, $v_5$},
    const plot,
    xlabel = {value},
  }
}

\colorlet{colora}{red!25!white}
\colorlet{colorb}{blue!25!white}
\colorlet{colorc}{yellow!25!white}
\colorlet{colord}{green!25!white}
\colorlet{colore}{orange!25!white}
\colorlet{colorl}{red!75!black}

\pgfplotsset{soldot/.style={color=blue,only marks,mark=*}}
\pgfplotsset{holdot/.style={color=blue,fill=white,only marks,mark=*}}

\pgfmathdeclarefunction{normpdf}{3}{%
\pgfmathparse{1/(#3*sqrt(2*pi))*exp(-((#1-#2)^2)/(2*#3^2))}%
}

\tikzset{declare function={
    normcdf(\x,\m,\s)=1/(1 + exp(-0.07056*((\x-\m)/\s)^3 - 1.5976*(\x-\m)/\s));
  }
}

\title{XPCA: Extending PCA for a Combination\\of Discrete and Continuous Variables}

\author{\name Clifford Anderson-Bergman\footnotemark[1] \email andersonberg1@llnl.gov\\
  \addr Lawrence Livermore National Laboratory\\
  \addr 7000 East Avenue\\
  \addr Livermore, CA 94551-0808, USA\\
  \\
  \name Tamara G. Kolda \email tgkolda@sandia.gov\\
  \name Kina Kincher-Winoto\footnotemark[2] \email kwinoto@sandia.gov\\
  \addr Sandia National Laboratories\\
  \addr 7011 East Avenue\\
  \addr Livermore, CA 94551-0969, USA}

\begin{document}

\jmlrheading{1}{2000}{1-48}{4/00}{10/00}{XPCA}{Clifford Anderson-Bergman, Tamara G. Kolda and Kina Kincher-Winoto} %
\firstpageno{1}
\ShortHeadings{XPCA: PCA for Continuous and Discrete Variables}{C. Anderson-Bergman, T. G. Kolda, K. Kincher-Winoto}
\editor{TBD}

\maketitle
\renewcommand*{\thefootnote}{(\fnsymbol{footnote})}
\footnotetext[1]{The work of this author was completed while he was employed at Sandia National Laboratories.}
\footnotetext[2]{Corresponding author.}
\renewcommand*{\thefootnote}{\arabic{footnote}.}

\begin{abstract}

Principal component analysis (PCA) is arguably the most popular tool
in multivariate exploratory data analysis. 
In this paper, we consider the question of how to handle heterogeneous variables that
include continuous, binary, and ordinal.
In the probabilistic interpretation of low-rank PCA,
the data has a normal multivariate distribution 
and, therefore, normal marginal distributions for each column.
If some  marginals are continuous but not normal, the
semiparametric copula-based principal component
analysis (COCA) method is an alternative to PCA that combines a
Gaussian copula with nonparametric marginals.
If some marginals are discrete or semi-continuous,
we propose a new extended PCA (XPCA) method
that also uses a Gaussian copula and nonparametric marginals
and accounts
for discrete variables in the likelihood calculation by integrating over appropriate intervals.
Like PCA, the factors produced by XPCA can be used to find latent structure in data, build predictive models, and 
perform dimensionality reduction. We present the new model, its induced likelihood function, and a fitting algorithm which 
can be applied in the presence of missing data.
We demonstrate how to use
XPCA to produce an estimated full conditional distribution for each data point, 
and use this to produce to provide estimates for missing data
that are automatically range respecting.
We compare the methods as applied to simulated and real-world data sets
that have a mixture of discrete and continuous variables.
\end{abstract}

\begin{keywords}
 principal component analysis (PCA), copula
 component analysis (COCA), matrix completion,
 heterogeneous data, matrix decomposition
\end{keywords}

\section{Introduction}
\label{sec:introduction}

Principal component analysis (PCA) is arguably the most popular tool
in multivariate exploratory data analysis. It can be
used to find latent structure in data, build predictive models,
perform dimensionality reduction, and more \citep{Jo02,AbWi10}.
Given an $m \times n$ data matrix $\X$, representing $n$ variables for
each of $m$ objects,
one use of PCA is to compute a \emph{low-rank approximation} of the form $\X \approx
\U\V'$. Here, $\U$ of size $m \times k$ is the matrix of \emph{factor scores} or \emph{loadings}, 
$\V$ of size $n \times k$ is the orthogonal matrix of \emph{principal components}, and $k
\ll m,n$.
In this paper, we consider the question of how to handle variables that
are not continuous. Specifically, we are interested in  binary and discrete ordinal variables
which have discontinuous
cumulative distribution functions~(CDFs).

To tackle this problem, we appeal to the probabilistic interpretation
of PCA per \citet{TiBi99} so that we can interpret the $\U$ and $\V$ matrices
in PCA as maximum likelihood estimators (MLEs).
In this interpretation, the data has a normal multivariate distribution, 
which implies normal marginal distributions for each column.
We describe this approach in \cref{sec:pca}.

Given that PCA can be viewed as a maximum likelihood estimator under the 
assumption of multivariate normality, several proposals have been made
to account for deviations from this assumption.
\citet{CoDaSc02} assume
the entries of a data matrix follow a distribution for the exponential family, 
and that the natural parameters can be estimated using low-rank matrix factorizations.
\citet{UdHoZaBo16} expand this idea to allow separate distributions 
for each column of the data, leading to column-specific loss functions that are 
scaled to balance their contributions to the overall loss.
Similarly, factor analysis methods model non-Gaussian random variables
through the use of generalized linear latent variable models (GLLVMs) as covered in depth in \citet{factorBook}.
The methods we present differ from these in that there is no need to
declare parametric distributions for each column of data. 

Other non-likelihood based methods have been adopted for non-Gaussian data. 
For binary and categorical data, multiple correspondence analysis (MCA) is a very popular method. 
An excellent review can be found in \citet{MCA}. In short, MCA expands categorical variables into binary variables, 
preprocess the data and then performs PCA. Similarly, Optimal Scaling, as presented in \citet{optimalScaling},
is an iterative method for ordinal, categorical or interval data, 
in which the  values are replaced with ordered values that optimize the variance explained. 
An excellent review can be found in \citet{optimalScalingReview}. 
While commonly used, these methods lack an underlying statistical model.
Recent work in \citet{probMCA} has proposed a statistical model for an alternative to MCA
by proposing a new estimator. In contrast, \citet{TiBi99} provided 
were able to develop a probabilistic model that justified the already established PCA estimator.
In this work, we will focus on likelihood-based methods in the framework established by \citet{TiBi99}. 

In the case that the marginals are continuous but not normal (e.g., heavily skewed), 
we can appeal to the theory of \emph{copulas} \citep{Sk73,Nelson06}
to separate the marginal distributions and their interactions.
We describe copulas in more detail in \cref{sec:copulas}.
\citet{HaLi12} propose
a semiparametric copula-based principal component
analysis (COCA) that
separates the estimation of the variable interactions and their marginals.
Specifically, the interactions are still assumed to be multivariate normal so that PCA can be used on the copula, but the marginals
can be any continuous distribution.
COCA is asymptotically equivalent to PCA when the marginals are normal.
The general idea of using semiparametric copulas can be found in 
\citet{GeGhRi95} in the sense that the marginal distributions are nonparametric
and the copulas are parametric.
\citet{EgKaScRo16} show the power of COCA in the context
of image analysis.
COCA is ingenious in its separation of the
marginals from the interactions so that PCA can be applied, but it depends on a translation of
each variable using its empirical distribution function.
We review empirical distribution functions in \cref{sec:columnw-marg-distr}.
and COCA in \cref{sec:coca}.
This transformation is
well-behaved for continuous data \emph{but not for discrete}~\citep{Ho07}.

\begin{figure}[t]
  \centering
  \subfloat[Normal]{\label{fig:ex-normal}
    \begin{tikzpicture}[scale=0.9]
      \begin{axis}[
        exdist,   
        ymin=0,ymax=1,
        xmin=-4,xmax=4,
        ]
        \addplot[domain=-4:4,samples=100]{normcdf(x,0,1)};      
        \node[] at (axis cs:2,.25) {\MethodList};
      \end{axis}
    \end{tikzpicture}          
    \hspace{-1.5mm}%
  }%
  \subfloat[Power law]{\label{fig:ex-powerlaw}
    \begin{tikzpicture}[scale=0.9]
      \begin{axis}[
        exdist,   
        ymin=0,ymax=1,
        xmin=1,xmax=10,
        ]
        \addplot[domain=1:10,samples=100]{1-(\x/1)^(-1.5)};      
        \node[] at (axis cs:7,.25) {\MethodList*};
      \end{axis}
    \end{tikzpicture}
  }%
  \subfloat[Zero-Inflated]{\label{fig:ex-zero-inflated}
    \begin{tikzpicture}[scale=0.9]
      \begin{axis}[
        exdist,   
        ymin=0,ymax=1,
        xmin=1,xmax=9,
        ]
        \addplot[domain=1:9,samples=100]{normcdf(x-5,0,1)*0.7+0.3};      
        \addplot[soldot] coordinates{(1,0.3)};
        \node[] at (axis cs:6.5,.25) {\MethodList**};
      \end{axis}
    \end{tikzpicture}
  }  
  \subfloat[Discrete]{\label{fig:ex-discrete2}
    \begin{tikzpicture}[scale=0.9]
      \begin{axis}[
        exdist,   
        const plot,
        ymin=0,ymax=1.0,
        xmin=-1,xmax=2,
        ]
        \draw[blue,very thick] (axis cs:-1,0)--(axis cs:0,0);
        \draw[blue,very thick, dotted] (axis cs:0,0)--(axis cs:0,0.75);
        \draw[blue,very thick] (axis cs:0,0.75)--(axis cs:1,0.75);
        \draw[blue,very thick, dotted] (axis cs:1,0.75)--(axis cs:1,1);
        \draw[blue,very thick] (axis cs:1,1)--(axis cs:2,1);
        \addplot[soldot] coordinates{(0,0.75) (1,1)};
        \addplot[holdot] coordinates{(0,0) (1,0.75)};
        \node[] at (axis cs:1,.25) {\MethodList**};
      \end{axis}
    \end{tikzpicture}          
  }  
  \caption{Differentiating the types of variables appropriate for each method.
    PCA works well for normal data, COCA works well for any continuous distribution, and
    XPCA works well for those plus semi-continuous (e.g., zero-inflated) and discrete data.}
  \label{fig:dists}
\end{figure}
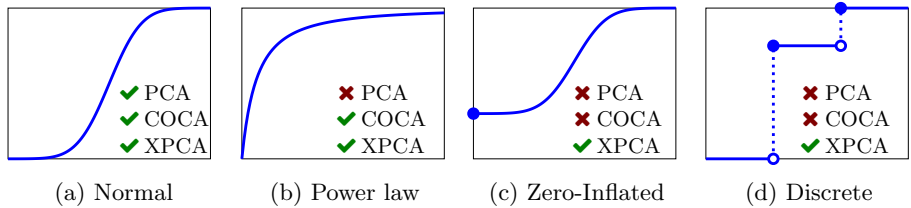

In \cref{sec:xpca}, we present a new method we call extended PCA (XPCA) for
the scenario in which one or more variables is discrete (or even semi-continuous).
Like COCA, we use a Gaussian copula and nonparametric marginals.
The difference is that we integrate over appropriate intervals in the likelihood to account
for discrete variables and
correctly compute the MLE.%
\footnote{The idea of integrating over an interval in the context of
  discrete variables is used, e.g., in probit ordinal regression
  \cite[Chapter~15.10]{Wo02}.}
XPCA is asymptotically equivalent to COCA for continuous variables
but handles discrete variables in a more appropriate manner
and yields improved performance on real-world
analysis tasks.
PCA, COCA, and XPCA are similar in that they all assume a Gaussian copula.
The difference is in the assumptions on the marginals, as illustrated  in \cref{fig:dists}.
PCA expects normal marginals, COCA expects continuous marginals, and XPCA can
handle general marginals including semi-continuous and discrete.

We
present methods for computing the XPCA model and also for inferring estimated values from it in \cref{sec:mapp-copula-pred}.
We demonstrate how one can derive a complete
conditional distribution for every entry of the data matrix as well as a more computationally efficient method for
estimating expected values.
These distributions can be used to identify
outliers, and the expected values can be used to fill in missing values.  
For binary variables, for example, the XPCA model can be used to generate the conditional probabilities for 0 and 1.
In contrast, COCA would simply return a 0 or 1 with no extra information, and
the value produced by PCA is hard to interpret because it could, for example, be negative or greater than 1.

In \cref{sec:examples}, we demonstrate the effectiveness of XPCA on both simulated and real data sets,
showing significant improvements as compared to PCA and COCA. We first compare the methods
on simulated data from different distributions,
showing how PCA or COCA may struggle when the assumptions are violated.
Then%
, we consider voting data from 18 years of the U.S.~Senate data, covering 271 senators and 9044 votes, with 63\% missing data
because not all senators were in office for all votes. We show how XPCA can be used to estimate how likely a particular vote might be.
We consider NBA player statistics and show that it does a better job at estimating missing entries than the other methods, 
captures relations between discrete variables missed by PCA and COCA, and present estimated 
distributions of data entries conditional on the fitted model. 

Our contributions may be summarized as follows:
\begin{itemize}[partopsep=0pt,topsep=0pt]
\item We propose the XPCA method, an extension to PCA/COCA that accounts for both continuous and discrete variables.
\item We develop an algorithm to fit the XPCA model to datasets, even
  in the context of incomplete data.
\item We show how the XPCA method can be used to derive a probability
  distribution for every entry of the data matrix. These distributions
  can in turn be used to identify outliers and infer missing values.
\item We demonstration the benefits of XPCA on simulated and real data sets, showing
  improvements as compared to both PCA and COCA.
\end{itemize}

\section{Background \& Related Work}
\label{sec:background}

\subsection{Notation}
\label{sec:notation}

We let $\Real$ denote the set of real values and
$\PosReal = (0,+\infty)$ denote the set of strictly positive real values.
We let $\Real^n$ denote the set of vectors of length $n$ and $\Real^{m \times n}$ denote the set of matrices of size $m \times n$.

We denote scalars as lowercase letters, vectors as boldface lowercase letters, and matrices as boldface uppercase letters.
The $j$th entry of a vector $\rowx$ is denoted by $x_j$. The $(i,j)$ entry of a matrix $\X$ is denoted as $\xij$.
We use the convention that the $i$th row of an $m \times n$ matrix $\X$ is denoted as $\rowxi$ and oriented as an $1 \times n$ object.
We let $\I$ denote the identity matrix whose size is determined by context.
We use the convention $\X \succ 0$ to denote the fact that $\X$ is symmetric positive definite.
We write $\rowx \leq \rowy$ if $x_j \leq y_j$ for all $j$.
We denote estimates with hats; e.g., $\X~$ denotes an estimator of $\X$.

Because we are using boldface, lowercase, and uppercase letters already to distinguish between scalars, vectors, and matrices, we instead
use \emph{color} to distinguish between random variables and their realizations. Hence, if $\RV{x}$ is a random variable, then
$x$ is its realization. We may write, for instance, $P(\RV{x} \leq x)$.

Throughout, we assume we are given a data matrix $\X$ of size $m \times n$ which is a realization
of some multivariate random variable $\RV\X$. 
The $m$ rows correspond to objects and the $n$ columns correspond to variables.
The data matrix $\X$ may have missing entries because not every variable is observed for every object; therefore, we provide some notation
to specify this. 
We let $\Omega_j$ denote the set of indices of known values in column $j$ so that
$(i,j) \in \Omega_j$ if and only if $\xij$ is observed.
We let $m_j$ denote the number of known entries in column $j$,
i.e., $m_j = | \Omega_j | \leq m$.
We define $\Omega = \bigcup_{j=1}^n \Omega_j$ to be the set of indices of all known entries.

We denote the univariate normal distribution with mean $\mu \in \Real$ and variance $\sigma^2 \in \PosReal$ as
$\Normal$. Its probability density function (PDF) is
\begin{equation}\label{eq:pdf-normal}
  f(x|\mu,\sigma^2) = \frac{ e^{(x-\mu)^2/2\sigma^2} }{\sqrt{2\pi\sigma^2}}.
\end{equation}
Since the distribution is continuous, the \emph{likelihood} of $\mu$ and $\sigma^2$ for a given
observation $x$ is simply the PDF, i.e., 
\begin{displaymath}
  \mathcal{L}(\mu,\sigma^2|x) = f(x|\mu,\sigma^2).
\end{displaymath}

We denote the multivariate normal distribution with mean $\bm{\mu} \in \Real^n$ and
covariance $\bm{\Sigma} \in \mathbb{\Real}^{n \times n}$ with $\bm{\Sigma} \succ 0$ as
$\MultiNormal[n]$. Its joint probability density function is
\begin{equation}\label{eq:mnpdf}
  f(\rowx | \bm{\mu}, \bm{\Sigma}) =  \det(2\pi \bm{\Sigma})^{-1/2}
  \exp\left( -\frac{1}{2} (\rowx - \bm{\mu})^{\transpose} \bm{\Sigma}^{-1} (\rowx - \bm{\mu}) \right).
\end{equation}
The likelihood of $\bm{\mu}$ and $\bm{\Sigma}$ for a given
observation $\rowx$ is again simply the PDF, i.e., 
\begin{displaymath}
  \mathcal{L}(\bm{\mu},\bm{\Sigma}|\rowx) = f(\rowx|\bm{\mu},\bm{\Sigma}).
\end{displaymath}
If $\bm{\Sigma} = \sigma^2 \I$, then the variables are independent and the likelihood reduces to
\begin{displaymath}
  \mathcal{L}(\bm{\mu},\bm{\Sigma}|\rowx) = \prod_{j=1}^n f(x_j|\mu_j,\sigma^2).
\end{displaymath}
We  slightly abuse notation and write
$\X* \sim \MultiNormal[m \times n]{\T}{\sigma^2 \I}$
to denote $\xij* \sim \Normal{\tij}{\sigma^2}$ for all $i=1,\dots,m$ and $j=1,\dots,n$.

Let $\phi: \Real \rightarrow \PosReal$ and $\Phi: \Real \rightarrow (0,1)$ denote the probability density
and cumulative distribution functions, respectively, for the univariate standard normal, i.e.,
\begin{equation}
  \label{eq:phi}
  \phi(x) = \frac{e^{-x^2/2}}{\sqrt{2\pi}}
  \qtext{and}
  \Phi(x) = \int_{-\infty}^x \phi(t) dt.
\end{equation}
We also use the  inverse CDF: $\PhiInv: [0,1] \rightarrow \Real \cup \set{-\infty,+\infty}$.
Writing things in terms of $\Phi$ and $\PhiInv$ is convenient for implementations since
routines exist to evaluate these quantities.

\subsection{PCA}
\label{sec:pca}

PCA is a standard technique for data analysis; see, e.g., \citet{Jo02} and \citet{AbWi10}.
\citet{TiBi99} provide a \emph{probabilistic} interpretation that is more 
conducive to the low-rank matrix factorization viewpoint and is our focus here.

We assume that we are given $m$ objects, each with $n$ variables.
Statistically, we represent this as a random variable $\X* \in \Real^{m \times n}$.
We standardize each column to define the random variable $\Z*$, i.e.,
\begin{equation}\label{eq:standardize}
  \zij* = \frac{\xij* - \hat \mu_j}{\hat \upsilon_j},
\end{equation}
where $\hat \mu_j$ and $\hat \upsilon_j$ are the estimated marginal mean and standard deviation, respectively,
for the random variables in the $j$th column of $\X*$.
This is equivalent to using the correlation matrix instead of the covariance matrix, 
which is generally recommended, as in \citet{Jo02}.
As the notation suggestions, $\zij*$ corresponds to the \emph{z-score} of $\xij*$, i.e., the number
of standard deviations that $\xij*$ is from its mean $\mu_j$.

In \emph{probabilistic PCA},  \citet{TiBi99} assume  $\Z*$ is such that 
each row $\rowz* \in \Real^{1 \times n}$ is distributed as
\begin{equation}\label{eq:zdist}
  \rowz* | \rowu* \sim \MultiNormal[n]{\rowu* \V'}{\sigma^2 \I}
  \qtext{with}
  \V \in \Real^{n \times k},
  \rowu* \sim \MultiNormal[k]{\bm{0}}{\I}.
\end{equation}
Here, $\V$ is the matrix of \emph{principal components}, i.e., combinations of
variables that together describe some aspect of the data and thus contribute to the correlations.
We have a dependence on the
random variable $\rowu*$, which specifies the combination of the principal
components. Finally, the $\sigma^2$ captures the variance \emph{not} described by the combination of principal components.%
\footnote{\citet{TiBi99} allow for nonzero marginal means, but we assume they are all zero thanks
  to the standardization.}
\citet{TiBi99} show that we can integrate out the $\rowu*$ in \cref{eq:zdist} to derive
\begin{equation}\label{eq:zdistalt}
  \rowz* \sim \MultiNormal[n]{\bm{0}}{\bm{\Sigma}}
  \qtext{where}
  \bm{\Sigma} = \V\V' + \sigma^2 \I.
\end{equation}
Then it is possible to work with $\Z'\Z$ to obtain estimates $\V~$ and $\hat\sigma$.
One difference between standard and probabilistic PCA is
that the $\V\V^\transpose$ matrix is low rank ($k < n$). If $\V \V^\transpose$ is full rank,
then the variance term $\sigma$ is redundant and can set to zero.

We take a slightly different tact in our approach, returning to \cref{eq:zdist}
and expressing it in matrix form as
\begin{equation}
  \label{eq:Zdist}
  \Z*|\U* \sim \MultiNormal[m \times n]{\U*\V'}{\sigma^2 \I}
  \qtext{where}
  \U* \sim \MultiNormal[m \times k]{\bm{0}}{\I}.
\end{equation}
Conditional on $\U$, if we define $\T=\U\V'$,
then each entry of $\Z*$ is independent with constant
variance, i.e.,
\begin{displaymath}
  \zij* \sim \Normal{\tij}.
\end{displaymath}
From \cref{eq:pdf-normal}, we can write the negative log-likelihood of the entire set of observations as 
\begin{equation}\label{eq:nll}
  \negloglik(\T,\sigma|\Z) =
  \sum_{(i,j) \in \Omega} \left( \frac{(\zij - \tij)^2}{2\sigma^2} + \log(\sigma)\right) + c
\end{equation}
where $c$ is a constant term.
We can minimize this to find the MLEs for $\U$ and $\V$.
The optimal values of $\U$ and $\V$ are equivalent for all values of $\sigma > 0$, so 
$\sigma$ can be ignored while finding $\U~$ and $\V~$.
As such, the MLEs of $\U$ and $\V$ can be obtained by minimizing the standard sum of squared errors:
\begin{equation}
  \label{eq:nllalt}
  \text{SSE}(\T|\Z) = 
  \sum_{(i,j) \in \Omega} (\zij - \tij)^2.  
\end{equation}
Using $\T~ =\U~\V~'$, then the MLE estimate of $\sigma$ is
\begin{displaymath}
  \hat\sigma = \sqrt{\sum_{(i,j) \in \Omega} (\zij-\tij~)^2 / | \Omega |}.
\end{displaymath}

A few caveats are in order. %
We typically assume that $\V$ is orthogonal \citep{AbWi10}.
If the results of the optimization are not orthogonal, then we can adjust by using the SVD factorization as follows:
\begin{equation}
  \label{eq:orthogonalize}
  \mathbf{\tilde U \tilde \Sigma \tilde V}^{\transpose} = \text{svd}(\U~\V~')
  \quad\Rightarrow\quad
  \U~ \gets \mathbf{\tilde U \tilde \Sigma}, \V~ \gets \mathbf{\tilde V}.
\end{equation}
If there are no missing entries, then we can use the SVD to compute the best rank-$k$ factorization
of $\X$ as follows:
Set $\U~$ to be the $k$ leading left singular values of $\X$ \emph{multiplied by the $k$ corresponding singular values},
and set $\V~$ to be the $k$ leading right singular values.
If some data is missing, then we need to apply optimization directly to \cref{eq:nllalt}.
The PCA method is given in \cref{alg:pca}.
We can use the results of the PCA to infer values as shown in \cref{alg:pca-impute}, and these estimates can be used
to for cross-validation, estimating  missing values, or detecting outliers.

\begin{algorithm}[th]
  \caption{PCA}\label{alg:pca}
  Let $\X$ be a data matrix of size $m \times n$ and $\Omega_j \subseteq \set{1,\dots,m}$ the known entries in column $j$.
  \begin{algorithmic}[1]
    \For{$j=1,\dots,n$}
    \State{$m_j \gets | \Omega_j |$}
    \Comment{Number of known entries in column $j$}
    \State{$\hat\mu_j \gets \frac{1}{m_j} \sum_{j \in \Omega_j} \xij$}
    \Comment{Compute marginal mean}
    \State{$\hat \upsilon_j \gets \sqrt{ \frac{1}{m_j} \sum_{j \in \Omega_j} (\xij - \hat\mu_j)^2}$}
    \Comment{Compute marginal standard deviation}
    \EndFor
    \State{$\Omega \gets \set{ (i,j) | i \in \Omega_j, j \in \set{1,\dots,n}}$}
    \Comment{Set of all known entries}
    \For {$(i,j) \in \Omega$} 
    \State{$\zij \gets ({\xij - \hat \mu_j})/{\hat \upsilon_j}$}
    \Comment{Standardize each column; \cref{eq:standardize}}
    \EndFor
    \State{$(\U~,\V~) \gets \arg\min \sum_{(i,j) \in \Omega} (z_{ij}-\theta_{ij})^2$  subject to $\T=\U\V'$}
    \Comment{MLE; see \cref{eq:nllalt}}
    \State{If necessary, adjust $(\U~,\V~)$ so that $\V~$ is orthogonal}
    \Comment{See \cref{eq:orthogonalize}}
  \end{algorithmic}
\end{algorithm}

\begin{algorithm}[th]
  \caption{PCA Impute}\label{alg:pca-impute}
  Let $\mathcal{S}$ denote the subset of entries to be imputed, 
  and let $\U~,\V~$ be the PCA factors.
  \begin{algorithmic}[1]
    \For {$(i,j) \in \mathcal{S}$} 
    \State{$\tij~ \gets \sum_{\ell=1}^k \hat u_{i\ell} \hat v_{j \ell}$} \Comment{Calculate single entry of $\T~=\U~\V~'$}
    \State{$\xij~ \gets \tij~ \hat \upsilon_j + \hat \mu_j$} \Comment Return to original scale
    \EndFor
  \end{algorithmic}
\end{algorithm}

\subsection{Copulas}
\label{sec:copulas}

Any $n$-dimensional distribution can be expressed as a composite object:
the marginal one-dimensional distribution for each of the $n$ variables and
an $n$-dimensional copula,  i.e.,
an $n$-dimensional distribution whose marginals are uniform on $[0,1]$ \citep{Nelson06}.
In this way, the marginals and their interactions
can be handled separately.
Specifically, Sklar's famous result \citep{Sk73} says that any
multivariate CDF $F$ can be expressed in terms of a copula $C$
as
\begin{displaymath}
  F(\rowx) = C(\rowy) \qtext{where} y_j = F_j(x_j) \text{ for } j=1,\dots,n.
\end{displaymath}
Here $F_j$ represents the continuous marginal distribution of the $j$th variate.
The advantage of the copula model is that this breaks up the model into two parts:
the marginal distributions given by $F_j$ and the relationships between the relative percentiles given by $C$. 

There are many types of copulas. We are interested specifically in
the case that $C$ is a \emph{Gaussian copula}, which means
\begin{equation}\label{eq:gaussian_copula}
  C(\rowy) = \int_{-\infty}^{\rowz} f(\rowt | \bm{\mu}, \bm{\Sigma}) d\rowt
  \qtext{where} z_j = \PhiInv(y_j) \text{ for } j=1,\dots,n,
\end{equation}
where $f$ is the multivariate normal PDF defined
in \cref{eq:mnpdf}.
Since each $\RV{y_j}$ is a standard uniform random variable, it must be the case
that $\rowz* \sim \MultiNormal$.

We can consider PCA from a copula viewpoint.
We are given a data matrix $\X$ of size $m \times n$.
We assume the marginals are Gaussian and map these to uniform marginals via the transformation:
\begin{displaymath}
   \yij* = \Phi\left( \frac{\xij* - \mu_j}{\upsilon_j} \right),
\end{displaymath}
where $\mu_j$ and $\upsilon_j$ are the mean and standard deviation of the $j$th marginal.
We assume a Gaussian copula, which means that
\begin{displaymath}
  \Z* \sim \MultiNormal{\T}{\sigma^2\I}
  \qtext{with} \zij* = \PhiInv(\yij*).
\end{displaymath}
We note that this shows that the standard practice of normalizing each column can be seen as
performing PCA with a Gaussian copula and Gaussian marginals.
Of course, the $\yij*$ is not explicit in PCA which is why the copula interpretation may not be immediately obvious.
It is worth noting that there are many other choices of copulas other than the multivariate Gaussian (see, e.g., \citet{copulas}). 
In general it is not straightforward how one should pick a given copula model, as discussed in \citet{Em09}. 
The Gaussian copula has been selected for two reasons: computational simplicity and 
to closer match the assumption of a probabilistic PCA model. 

We subsequently describe COCA and our new XPCA, both of which also consider Gaussian copulas but have different assumptions about the marginals.

\subsection{Empirical Distribution Function}
\label{sec:columnw-marg-distr}

To estimate the marginal distribution, 
we use the nonparametric \emph{empirical distribution function} (EDF).
Since we are factorizing a matrix $\X$ of size $m \times n$ where each column has
its own distribution, we explain the EDF in that context.

For column $j$, we assume the random variables $\RV{\xij}$ are
drawn from an unknown distribution whose marginal CDF is
$F_j: \Real \rightarrow [0,1]$. 
Note that this requires that the data be ordered (as opposed to categorical), so continuous, semi-continuous, and discrete values are allowed.

Because some of data is discrete or semi-continuous, we are likely to have repeated observations.
Hence, we develop some machinery to cope with that.
Let the set of \emph{distinct} entries in column $j$ be
\begin{equation}\label{eq:Cj}
  \Cj = \set{ \xij | (i,j) \in \Omega_j}.
\end{equation}
If some values in the column are repeated, then $|\Cj| < m_j$,
where $m_j \leq m$ is the total number of entries (including repeats but not including missing entries) in column $j$.

We present two common methods for defining the EDF, both of which
are relevant for our analysis and comparison of the methods. 
The first and more common way to define the EDF is denoted here by $\Fhat_j$ and is given by
\begin{equation}\label{eq:edf}
  \Fhat_j(x) = \frac{\max \set{r_j(s) | s \leq x \text{ and } s \in \Cj}} {m_j},  
\end{equation}
where $r_j(s) \in \set{1,...,m_j}$ is the rank of $s$
relative to the data in column $j$, and the \emph{maximum} is used in the case of ties. 
If $x \leq s$ for all $s \in \Cj$, then $\Fhat_j(x) = 0$.
Then the inverse is given by
\begin{equation}\label{eq:edfinv}
  \FhatInv_j(y) = 
  \begin{cases}
    \min \set{s \in \Cj} & \text{if } y \leq \min \set{ \Fhat_j(s) | s \in \Cj }, \\
    \max \set{s \in \Cj | \Fhat_j(s) \leq y} & \text{otherwise}.   
  \end{cases}
\end{equation}
Observe $\FhatInv_j:[0,1]\rightarrow \Cj$.
The second way to define the EDF, which we will use for COCA, is denoted here as
$\Falt$ and is given by
\begin{equation}\label{eq:edf-alt}
  \Falt_j(x) =  \frac{\max \set{\tilde r_j(s) | s \leq x \text{ and } s \in \Cj}} {m_j + 1},
\end{equation}
where $\tilde r_j(s) \in \set{ 1,..., m_j}$ is the rank of $s$
relative to the data in column $j$, and the \emph{midpoint} is used in the case of ties.
 As with $\Fhat$, if $x \leq s$ for all $s \in \Cj$, then $\Falt_j(x) = 0$.
Likewise, we let $\FaltInv_j:[0,1] \rightarrow \Cj$ be defined analogously, replacing $\Fhat_j$ with
$\Falt_j$.
In both cases, the inverse maps to the set of observed values, $\Cj$.

There are two differences between $\Fhat_j$ and $\Falt_j$.
The first difference is the denominator: $m_j$ versus $m_j+1$.
This modification is not so unusual and can be found, for instance, in \citet[p.~36, Definition 2.4]{CoBaTrDo01}.
Using $m_j+1$ ensures
that $\Falt(x) \in (0,1)$ for all $x \in \Cj$, which is critical for COCA for reasons we explain 
in \cref{sec:coca}.
The second difference is using the maximum versus the midpoint for breaking ties.
This is an ad hoc way for COCA to handle discrete variables since, in our experiments, mapping
to the midpoint yielded better results than mapping to the maximum (see \cref{sec:midp-imput-coca}).  

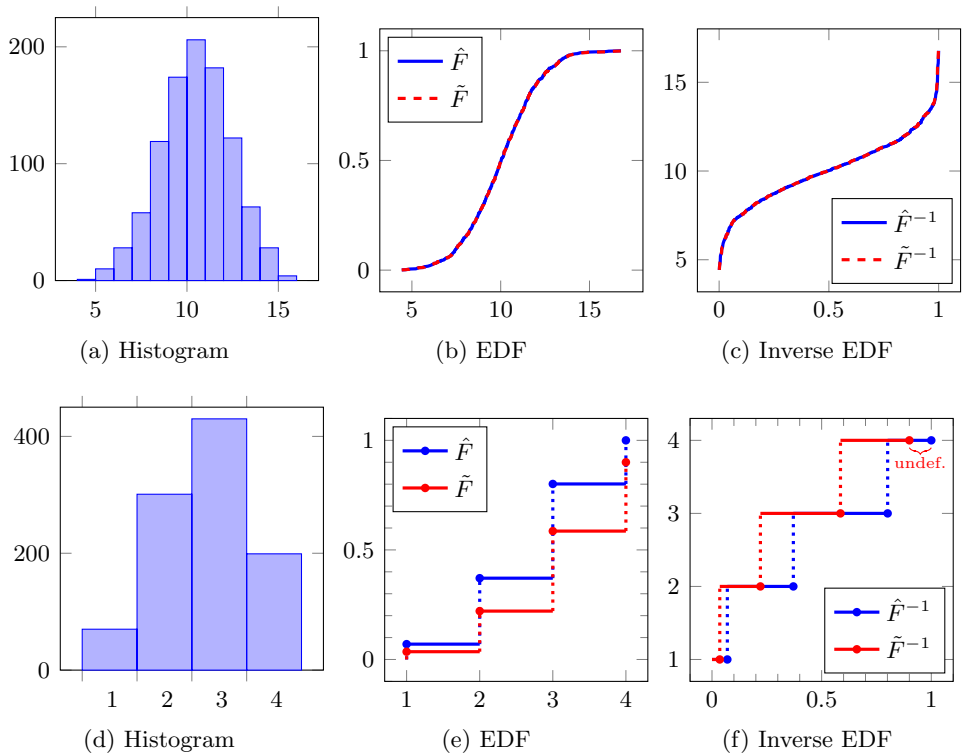
\begin{figure}[t]
  \centering
  \subfloat[Histogram]{
    \begin{tikzpicture}
      \begin{axis}[ybar,ymin=0, ymax=225]
        \addplot+[ybar interval,mark=no] plot coordinates { (4, 1) (5, 10) (6, 28) (7, 58) (8, 119)
          (9, 174) (10, 206) (11, 182) (12, 122) (13, 63) (14, 28) (15, 4) (16, 5) 
        };
      \end{axis}
    \end{tikzpicture}}
  \subfloat[EDF]{
    \begin{tikzpicture}
      \begin{axis}[width=2in, height=2in,
        legend pos=north west,
        legend entries={$\hat F$, $\tilde F$},
        legend style={font=\footnotesize},
        ]
        \addplot[blue, very thick] table {edfdata-fhat.dat};
        \addplot[red, very thick, dashed] table {edfdata-ftilde.dat};
      \end{axis}  
    \end{tikzpicture}}
  \subfloat[Inverse EDF]{
    \begin{tikzpicture}
      \begin{axis}[width=2in, height=2in,
        legend pos=south east,
        legend entries={$\hat F^{-1}$, $\tilde F^{-1}$},
        legend style={font=\footnotesize},
        ]
        \addplot[blue, very thick] table[x index={1}, y index={0}] {edfdata-fhat.dat};
        \addplot[red, very thick, dashed] table[x index={1}, y index={0}] {edfdata-ftilde.dat};
      \end{axis}  
    \end{tikzpicture}         
  }
  
  \subfloat[Histogram]{
    \begin{tikzpicture}
      \begin{axis}[ybar,ymin=0, ymax=450, xtick={1,2,3,4},xticklabel style = {xshift=+4mm}]
        \addplot+[ybar interval,mark=no] plot coordinates { (1, 70) (2, 301) (3, 430) (4, 199) (5,0) };
      \end{axis}
    \end{tikzpicture}}
  \subfloat[EDF]{
    \begin{tikzpicture}
      \begin{axis}[width=2in, height=2in,
        legend pos=north west,
        legend entries={$\hat F$, $\tilde F$},
        legend style={font=\footnotesize},
        minor y tick num = 4]
        \addplot[color=blue, mark=*,jump mark left,very thick,mark options={scale=0.5}] coordinates{ ( 1.000000, 0.070000 ) ( 2.000000, 0.371000 ) ( 3.000000, 0.801000 ) ( 4.000000, 1.000000 )  }; 
        \addplot[color=blue,forget plot,dotted,very thick] coordinates{ ( 1.000000, 0.000000 ) ( 1.000000, 0.070000 )  }; 
        \addplot[color=blue,forget plot,dotted,very thick] coordinates{ ( 2.000000, 0.070000 ) ( 2.000000, 0.371000 )  }; 
        \addplot[color=blue,forget plot,dotted,very thick] coordinates{ ( 3.000000, 0.371000 ) ( 3.000000, 0.801000 )  }; 
        \addplot[color=blue,forget plot,dotted,very thick] coordinates{ ( 4.000000, 0.801000 ) ( 4.000000, 1.000000 )  }; 
        \addplot[color=red, mark=*,jump mark left,very thick,mark options={scale=0.5}] coordinates{ ( 1.000000, 0.035465 ) ( 2.000000, 0.220779 ) ( 3.000000, 0.585914 ) ( 4.000000, 0.900100 )  }; 
        \addplot[color=red,forget plot,dotted,very thick] coordinates{ ( 1.000000, 0.000000 ) ( 1.000000, 0.035465 )  }; 
        \addplot[color=red,forget plot,dotted,very thick] coordinates{ ( 2.000000, 0.035465 ) ( 2.000000, 0.220779 )  }; 
        \addplot[color=red,forget plot,dotted,very thick] coordinates{ ( 3.000000, 0.220779 ) ( 3.000000, 0.585914 )  }; 
        \addplot[color=red,forget plot,dotted,very thick] coordinates{ ( 4.000000, 0.585914 ) ( 4.000000, 0.900100 )  }; 
      \end{axis}  
    \end{tikzpicture}}
  \subfloat[Inverse EDF]{
    \begin{tikzpicture}
      \begin{axis}[width=2in, height=2in,
        legend pos=south east,
        legend entries={$\hat F^{-1}$, $\tilde F^{-1}$},
        legend style={font=\footnotesize},
        minor x tick num = 4]
        \addplot[color=blue, mark=*,jump mark right,very thick,mark options={scale=0.5}] coordinates{ ( 0.070000, 1.000000 ) ( 0.371000, 2.000000 ) ( 0.801000, 3.000000 ) ( 1.000000, 4.000000 )  }; 
        \addplot[color=blue,forget plot,very thick] coordinates{ ( 0.000000, 1.000000 ) ( 0.070000, 1.000000 )  }; 
        \addplot[color=blue,forget plot,dotted,very thick] coordinates{ ( 0.070000, 1.000000 ) ( 0.070000, 2.000000 )  }; 
        \addplot[color=blue,forget plot,dotted,very thick] coordinates{ ( 0.371000, 2.000000 ) ( 0.371000, 3.000000 )  }; 
        \addplot[color=blue,forget plot,dotted,very thick] coordinates{ ( 0.801000, 3.000000 ) ( 0.801000, 4.000000 )  }; 
        \addplot[color=red, mark=*,jump mark right,very thick,mark options={scale=0.5}] coordinates{ ( 0.035465, 1.000000 ) ( 0.220779, 2.000000 ) ( 0.585914, 3.000000 ) ( 0.900100, 4.000000 )  };
        \draw[color=red,decorate,decoration={brace,mirror}] (axis cs:0.9,3.9) -- (axis cs:1,3.9);
        \node at (axis cs: 0.95,3.7) {\textcolor{red}{\tiny undef.}};
        \addplot[color=red,forget plot,very thick] coordinates{ ( 0.000000, 1.000000 ) ( 0.035465, 1.000000 )  }; 
        \addplot[color=red,forget plot,dotted,very thick] coordinates{ ( 0.035465, 1.000000 ) ( 0.035465, 2.000000 )  }; 
        \addplot[color=red,forget plot,dotted,very thick] coordinates{ ( 0.220779, 2.000000 ) ( 0.220779, 3.000000 )  }; 
        \addplot[color=red,forget plot,dotted,very thick] coordinates{ ( 0.585914, 3.000000 ) ( 0.585914, 4.000000 )  }; 
      \end{axis}  
    \end{tikzpicture}         
  }
  \caption{Comparison of the two definitions of the EDF, $\Fhat_j$ and $\Falt_j$. The data for the EDFs
    consists of $m_j=1000$ points drawn from a normal distribution with $\mu=10$ and $\sigma=2$ (top row) and discrete distribution with 4 values (bottom row).
    The differences between the two definitions are nominal for the continuous normal distribution and dramatic for the discrete one.}
  \label{fig:edf-compare}
\end{figure}

For continuous distributions, both $\Fhat_j$ and $\Falt_j$ 
are consistent nonparametric estimators of $F_j$. For discrete variables, 
only $\Fhat_j$ is a consistent estimator of $F_j$.
A comparison of the two definitions of the EDF is shown in \cref{fig:edf-compare}. In the continuous case, there
is little visual difference between the two definitions. In the discrete case, however, there are two major differences.
First, $\Falt$ never gets to 1.
Second, the use of the midpoint impacts the  y-values for $\Falt$.

\subsection{COCA}
\label{sec:coca}

In PCA as described in \cref{sec:pca}, one assumption of the method is that the marginals are assumed to be normal.
\citet{HaLi12} proposed COCA,
a semiparametric version of PCA, to relax this assumption.
The idea is to still use a parametric Gaussian copula, as in PCA, but to
use nonparametric EDFs for the marginals.
So, assume we are given a data matrix $\X$ of size $m \times n$.
We create variables with uniform marginal distribution via the transformation:
\begin{displaymath}
   \yij* = F_j( \xij*),
 \end{displaymath}
 where $F_j$ is the $j$th marginal EDF.
As with PCA, we assume a Gaussian copula, which means that
\begin{displaymath}
  \Z* \sim \MultiNormal{\T}{\sigma^2\I}
  \qtext{with} \zij* = \PhiInv(\yij*).
\end{displaymath}

\citet{EgKaScRo16} applied COCA in the context of face modeling.
They observed that the color intensity
distributions in images were not normally distributed but rather bimodal. 
They compared COCA and PCA
on image processing applications.

We use the version of COCA from \citet{EgKaScRo16} in the remainder of our discussion.
The procedure is given in \cref{alg:coca}.
In \cref{line:coca-edf},
it is vital that $\Falt_j$ be used
because it avoids an infinite $\zij$ value in \cref{line:coca-phiinv}.
If we used $\Fhat_j$ instead, those lines would produce
\begin{displaymath}
\yij = \Fhat_j\bigl( \max_i \{ \xij \} \bigr) = 1
\quad \Rightarrow \quad
\PhiInv (\yij) = \infty. 
\end{displaymath}
As such, using $\Fhat_j$ would result in infinite values in $\Z$, which would make factorizing it impossible.
\citet{GeGhRi95} makes the observation that ``this rescaling avoids difficulties arising from\dots potential unboundedness.''

\begin{algorithm}[t]
  \caption{COCA}\label{alg:coca}
Let $\X$ be a data matrix of size $m \times n$ and $\Omega_j \subseteq \set{1,\dots,m}$ the known entries in column $j$.
  \begin{algorithmic}[1]
    \For{$j=1,\dots,n$}
    \State{$\Falt_j \gets$ EDF for column $j$ per \cref{eq:edf-alt}}
    \Comment{Compute marginal distributions}
    \EndFor
    \For {$(i,j) \in \Omega$} 
    \State{\label{line:coca-edf}$\yij \gets \Falt_j(\xij)$}
    \Comment{Transform original marginals to standard uniform}
    \State{\label{line:coca-phiinv}$\zij \gets \PhiInv(\yij)$}
    \Comment{Transform standard uniform marginals to Gaussian}
    \EndFor
    \State
    $(\U~,\V~) \gets $ argmin of \cref{eq:nllalt} subject to $\T=\U\V'$    
    \State If necessary, adjust $(\U~,\V~)$ so that $\V~$ is orthogonal
    \Comment{See \cref{eq:orthogonalize}}
  \end{algorithmic}
\end{algorithm}

The $\hat \T$ estimates are then pushed through the corresponding $\FaltInv_j$'s to provide 
point estimates $\hat \xij$, as given in \cref{alg:coca-impute}.
In contrast to PCA, every estimate $\xij~$ must be an observed value from column $j$, i.e., $\xij~ \in \Cj$.
The benefit is that we can never, for instance, estimate a negative value when there have only been nonnegative observations.

\begin{algorithm}[t]
  \caption{COCA Impute}\label{alg:coca-impute}
    Let $\mathcal{S}$ denote the subset of entries to be estimated.
  \begin{algorithmic}[1]
    \For {$(i,j) \in \mathcal{S}$} 
    \State $\tij~ \gets \sum_{\ell=1}^k \hat u_{i\ell} \hat v_{j \ell}$ \Comment{Calculate single entry of $\T~=\U~\V~'$}
    \State $\yij~ \gets \Phi(\tij~)$ \Comment Transform standard normal to standard uniform
    \State $\xij~ \gets \FaltInv_j(\yij~)$ \Comment Transform standard uniform to original distribution
    \EndFor
  \end{algorithmic}
\end{algorithm}

\section{XPCA}
\label{sec:xpca}

We present our new method, XPCA, an extension to PCA and COCA that handles 
discrete variables. We present the generative model, 
the induced likelihood function, and two optimization approaches to 
find the maximum likelihood estimates.

\subsection{Model}
\label{sec:model}
    
Similar to COCA, we assume a Gaussian copula model that follows a PCA generative model. 
In other words, we assume there is a latent variable  $\Z*$ such that
\begin{displaymath}
  \Z* \sim \MultiNormal{\T}{\sigma^2\I} 
 \end{displaymath}
where $\T$ is a low-rank matrix. The values of $\Z*$ are not observed exactly but 
rather through the transformations 
\begin{displaymath}
\yij* = \Phi( \zij* )
\qtext{and}
\xij* = F^{-1}_j( \yij* ).
\end{displaymath}
While we assume that $\Z*$ comes from a multivariate Gaussian PCA
model, we make no assumptions about the form of $F_j$ other than
that it is a proper CDF. The key difference between XPCA and COCA is that COCA implicitly assumes that $F_j$ 
corresponds to a continuous random variable and so is itself continuous.
In contrast, XPCA allows for jump discontinuities in $F_j$, as discussed in the next subsection.

\subsection{Likelihood}
\label{sec:likelihood}

We are interested in the case when $F_j$ is the CDF of a 
discrete random variable. In such cases, $F_j$ is not 
a 1-to-1 function and indeed a \emph{range} of values of $\zij*$ (and of the corresponding $\yij*$) will all lead 
to the same $\xij*$. In these cases, \citet{Ho07} and  \citet{LiLiQiYa16}  suggest that we can still use
standard copulas, such as the Gaussian copula, to model joint distributions of discrete random variables,
but the likelihood must be computed over a range of $\yij*$ values to achieve consistent estimation. 
\citet{Ho07} was interested in examining correlation coefficients for 
ordinal data with a Gaussian copula, but the general principle of that work can
be applied in our context as well.

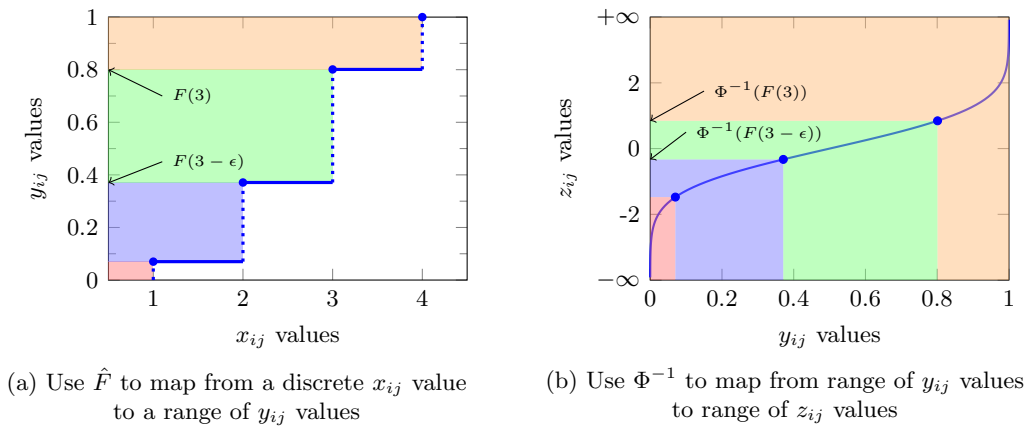
\begin{figure}[t]
  \centering
  \subfloat[Use $\hat F$ to map from a discrete $\xij$ value to a range of $\yij$ values]{
    \begin{tikzpicture}
      \begin{axis}[
        width=2.5in, height=2in,
        minor y tick num = 1,
        xmin = 0.5, xmax = 4.5,
        ymin = 0, ymax = 1,
        xlabel = $\xij$ values,
        ylabel = $\yij$ values,
        ]
\filldraw[fill=red!50,fill opacity=0.5,draw opacity=0] (axis cs: 1.000000, 0.000000 ) -- (axis cs: 1.000000, 0.070000 ) -- (axis cs: 0.500000, 0.070000 ) -- (axis cs: 0.500000, 0.000000 ) --  cycle; 
\filldraw[fill=blue!50,fill opacity=0.5,draw opacity=0] (axis cs: 2.000000, 0.070000 ) -- (axis cs: 2.000000, 0.371000 ) -- (axis cs: 0.500000, 0.371000 ) -- (axis cs: 0.500000, 0.070000 ) --  cycle; 
\filldraw[fill=green!50,fill opacity=0.5,draw opacity=0] (axis cs: 3.000000, 0.371000 ) -- (axis cs: 3.000000, 0.801000 ) -- (axis cs: 0.500000, 0.801000 ) -- (axis cs: 0.500000, 0.371000 ) --  cycle; 
\filldraw[fill=orange!50,fill opacity=0.5,draw opacity=0] (axis cs: 4.000000, 0.801000 ) -- (axis cs: 4.000000, 1.000000 ) -- (axis cs: 0.500000, 1.000000 ) -- (axis cs: 0.500000, 0.801000 ) --  cycle; 
\addplot[color=blue, mark=*,jump mark left,very thick,mark options={scale=0.5}] coordinates{ ( 1.000000, 0.070000 ) ( 2.000000, 0.371000 ) ( 3.000000, 0.801000 ) ( 4.000000, 1.000000 )  }; 
\addplot[color=blue,forget plot,dotted,very thick] coordinates{ ( 1.000000, 0.000000 ) ( 1.000000, 0.070000 )  }; 
\addplot[color=blue,forget plot,dotted,very thick] coordinates{ ( 2.000000, 0.070000 ) ( 2.000000, 0.371000 )  }; 
\addplot[color=blue,forget plot,dotted,very thick] coordinates{ ( 3.000000, 0.371000 ) ( 3.000000, 0.801000 )  }; 
\addplot[color=blue,forget plot,dotted,very thick] coordinates{ ( 4.000000, 0.801000 ) ( 4.000000, 1.000000 )  }; 
\draw[black,->] (axis cs: 1.1, 0.7) node[right] {\tiny $F(3)$} -- (axis cs: 0.5,0.8);
\draw[black,->] (axis cs: 1.1, 0.45) node[right] {\tiny $F(3-\epsilon)$} -- (axis cs: 0.5,0.37);
\end{axis}  
\end{tikzpicture}}
  ~~~~~~
  \subfloat[Use $\PhiInv$ to map from range of $\yij$ values to range of $\zij$ values]{
    \begin{tikzpicture}
      \begin{axis}[
        width=2.5in, height=2in,
        ymin = -4, ymax = 4,
        xmin = 0, xmax = 1,
        xlabel = $\yij$ values,
        ylabel = $\zij$ values,
        ytick = {-4,-2,0,2,4},
        yticklabels = {$-\infty$,-2,0,2,$+\infty$},
        ]
        \addplot[blue,thick] table {norminv.dat};
\filldraw[fill=red!50,fill opacity=0.5,draw opacity=0] (axis cs: 0.000000, -4.000000 ) -- (axis cs: 0.070000, -4.000000 ) -- (axis cs: 0.070000, -1.475791 ) -- (axis cs: 0.000000, -1.475791 ) -- (axis cs: 0.000000, -4.000000 ) -- (axis cs: 0.000000, -4.000000 ) --  cycle; 
\filldraw[fill=blue!50,fill opacity=0.5,draw opacity=0] (axis cs: 0.070000, -4.000000 ) -- (axis cs: 0.371000, -4.000000 ) -- (axis cs: 0.371000, -0.329206 ) -- (axis cs: 0.000000, -0.329206 ) -- (axis cs: 0.000000, -1.475791 ) -- (axis cs: 0.070000, -1.475791 ) --  cycle; 
\filldraw[fill=green!50,fill opacity=0.5,draw opacity=0] (axis cs: 0.371000, -4.000000 ) -- (axis cs: 0.801000, -4.000000 ) -- (axis cs: 0.801000, 0.845199 ) -- (axis cs: 0.000000, 0.845199 ) -- (axis cs: 0.000000, -0.329206 ) -- (axis cs: 0.371000, -0.329206 ) --  cycle; 
\filldraw[fill=orange!50,fill opacity=0.5,draw opacity=0] (axis cs: 0.801000, -4.000000 ) -- (axis cs: 1.000000, -4.000000 ) -- (axis cs: 1.000000, 4.000000 ) -- (axis cs: 0.000000, 4.000000 ) -- (axis cs: 0.000000, 0.845199 ) -- (axis cs: 0.801000, 0.845199 ) --  cycle; 
\addplot[color=blue,only marks, mark=*,mark options={scale=0.75}] coordinates{
(0.07,-1.47579) 
(0.371,-0.329206) 
(0.801,0.845199) 
};
\draw[black,->] (axis cs: 0.15, 1.75) node[right] {\tiny $\PhiInv(F(3))$} -- (axis cs: 0,0.845);
\draw[black,->] (axis cs: 0.1, 0.5) node[right] {\tiny $\PhiInv(F(3-\epsilon))$} -- (axis cs: 0,-0.33);
      \end{axis}
    \end{tikzpicture}
  }
  \caption{Illustrating how discrete $\xij$ values map to ranges of $\zij$ values. The value of $\xij=3$ is explicitly shown.}
  \label{fig:zij}
\end{figure}

Leveraging these ideas, we can see that if the interval $(\lij, \rij]$ is the range of values
of $\zij*$ such that $\xij* = \xij$ is observed, then 
\begin{displaymath}
P(\xij* = \xij) = \int_{\lij}^{\rij} f(t | \theta_{ij}, \sigma^2) dt =
\Phi \left( \frac{\rij - \theta_{ij}}{\sigma} \right) - \Phi \left( \frac{\lij - \theta_{ij}}{\sigma} \right),
\end{displaymath}
where $f$ is the PDF of a normal distribution from \cref{eq:pdf-normal}. 
The range of $\zij$ values corresponding to a given $\xij$ is given by
$[\PhiInv( F_j( \xij -\epsilon) ), \PhiInv( F_j( \xij) )]$ for sufficiently small~$\epsilon$.
This is illustrated in \cref{fig:zij}.
Hence, the negative log-likelihood with respect to discrete random variables
can be written as 
\begin{multline}\label{eq:nll-xpca}
  \negloglik(\T, \sigma, F | \X) = \sum_{(i,j) \in \Omega}
  \negthickspace\negthickspace
  - \log \left[ \Phi\left( \frac{\rij - \tij}{\sigma} \right) - \Phi\left( \frac{\lij - \tij}{\sigma} \right) \right] 
  \qtext{where} \\
 \lij = \PhiInv(F_j(\xij-\epsilon)) \qtext{and} \rij = \PhiInv(F_j(\xij)).
\end{multline} 
In practice, we set $\epsilon$ to be half the distance between the closest two distinct points in any column, i.e.,
\begin{equation}
  \label{eq:epsilon}
  \epsilon = \frac{1}{2} \min
  \set{\xi - \xi' | \xi, \xi' \in \Cj,\, \xi > \xi',\, j=1,\dots,n }.
\end{equation}

We observe that asymptotically, the XPCA likelihood reduces to the COCA likelihood plus a constant
if all the variables are continuous.
If $F_j$ has jump discontinuities (e.g., because it corresponds to a discrete variable),
then $\lij \not\rightarrow \rij$ as $\epsilon \downarrow 0$.
However, if $F_j$ is continuous (because it corresponds to a continuous variable),
then $\lij \rightarrow \rij$ as $\epsilon \downarrow 0$ and so we see an asymptotic equivalence to COCA because
\begin{displaymath}
\lim_{\lij \rightarrow \rij}    \frac{ \Phi\left( \frac{\rij - \tij}{\sigma} \right) - \Phi\left( \frac{\lij - \tij}{\sigma} \right)} 
{  ( \rij - \lij) } = f( \rij | \theta_{ij}, \sigma^2). 
\end{displaymath}
Hence, if $F_j$ corresponds to a continuous random variable, the contribution to 
the negative log-likelihood will be asymptotically equivalent, excepting the additive term 
$c_{ij} = \log(\rij - \lij)$. The $c_{ij}$ has no dependence on the parameters $\U, \V, \sigma$
and so is irrelevant for the maximum likelihood estimate.
As such, the XPCA solution will approach the COCA 
solution if all the data is continuous as we demonstrate on simulated data in \cref{sec:simulated-examples}.
A comparison of the different methods is shown in \cref{fig:comparison}.

\begin{figure}[th]
  \centering\small
  \begin{tabular}{|cc|c|c|}
    \hline
    & PCA & COCA & XPCA \\[5mm]
    \em Transformation
          & $\zij* = (\xij*-\mu_j)/{\nu_j}$
          & $\zij* = \PhiInv(F_j(\xij*))$
          &  $\zij* = \PhiInv(F_j(\xij*))$  \\[5mm]
    \em Copula
          & $\Z* \sim \MultiNormal{\T}{\sigma^2\I}$
          & $\Z* \sim \MultiNormal{\T}{\sigma^2\I}$
          & $\Z* \sim \MultiNormal{\T}{\sigma^2\I}$  \\[5mm]
    \em Z-value
          & $\zij = (\xij - \hat\mu_j)/\hat \upsilon_j$
          & $\zij = \PhiInv(\Falt_j(\xij))$
                 & $\zij \in (\;  \underbrace{\PhiInv(\Fhat_j(\xij{-}\epsilon))}_{\lij}, \underbrace{\PhiInv(\Fhat_j(\xij))}_{\rij}
                   \; ]$ \\[8mm]
    \em Effective Loss 
          & $(\zij-\tij)^2$
          & $(\zij-\tij)^2$
          & ${-}\log \left[
                \Phi\!\left( \frac{\rij {-} \tij}{\sigma} \right) -
                \Phi\!\left( \frac{\lij {-} \tij}{\sigma} \right)
            \right]$
            \\ \hline
  \end{tabular}
  \caption{Comparison of the underlying statistical models for PCA, COCA, and XPCA. The top line is the assumed data transformation based on the marginal distributions, the second line is the copula (same for all three), the third line is the actual computation of the z-values that should be from a Gaussian distribution, and the last line is the effective loss from the negative log likelihood.}
  \label{fig:comparison}
\end{figure}

\subsection{Fitting the Model}
\label{sec:fitting-model} 
Like COCA, we use the EDFs to estimate $F_j$. Because 
the full interval $(\lij, \rij]$ is considered, $\Fhat_j$ can be used instead of $\Falt_j$.
The parameters $\U, \V$ and $\sigma$ 
are then estimated using maximum likelihood estimation. 
With standard PCA, it is not necessary to include $\sigma$ when 
solving for $\U~, \V~$, because
the maximum likelihood estimates of $\U$ and $\V$ do not depend on $\sigma$.
For XPCA, this is not the case and $\sigma$ must estimated concurrently 
with $\U$ and $\V$. 

Similar to most factorization problems, the negative log-likelihood is nonconvex
as a function of $\{ \U, \V, \sigma \}$, but convex as a function of $\U$ conditional 
on $\{ \V, \sigma \}$ and $\V$ conditional on $\{ \U, \sigma \}$. This suggests two 
straightforward optimization approaches:
\begin{itemize}
\item All-at-once optimization, i.e., using a quasi-Newton method, or
\item Alternating optimization, i.e., block coordinate descent.
\end{itemize}

All-at-once optimization is inviting due to the simplicity of fewer hyperparameters and general speed.
For this. we use L-BFGS \citep{No80,ByLuNoZh95} as implemented in NLopt \citep{NLOPT}.

Because we found that L-BFGS would occasionally have line-search failures,
we also implemented a block coordinate descent algorithm
that was more robust at the cost of longer computation times. 
Not only 
is the loss function convex in $\U$,
but the loss is independent for each row $\U$ (and likewise for $\V$). 
Thus, each iteration of our block coordinate descent algorithm updates 
each row of $\U$ and $\V$ independently, and these updates use a single step of Newton-Raphson. 
After $\U$ and $\V$ have been updated, $\sigma$ is updated. If the second derivative 
with respect to $\sigma$ is positive, Newton's method with half-stepping
is used to update $\sigma$; otherwise, gradient descent with a line search is used. 

These approached requires the gradient and  Hessian,
whose formulas are given in \cref{sec:derivatives}. 

\subsection{Algorithm}
\label{sec:algorithm}

The XPCA method is presented in \cref{alg:xpca}. The method computes the EDFs based on the observed data. These functions are used to compute the ranges for possible z-values, which corresponds to the integration limits for the loss function. The optimization of the loss function is computed as described in \cref{sec:fitting-model}. We include a final step of adjusting the factors so that $\V~$ is orthogonal, but this is optional.

\begin{algorithm}[t]
  \caption{XPCA}\label{alg:xpca}
  Let $\X$ be a data matrix of size $m \times n$ and
  $\Omega_j \subseteq \set{1,\dots,m}$ the known entries in column $j$.
  \begin{algorithmic}[1]
    \For{$j=1,\dots,n$}
    \State $\Fhat_j \gets$ EDF for column $j$ per \cref{eq:edf}
    \Comment Compute marginal distributions
    \EndFor
    \State $\epsilon \gets
    \frac{1}{2} \min
    \set{\xi - \xi' | \xi, \xi' \in \Cj,\, \xi > \xi',\, j=1,\dots,n }$
    \For {$(i,j) \in \Omega$} 
    \State $\lij \gets \PhiInv( \Fhat(\xij - \epsilon) )$ \Comment Lower bound of z-range
    \State $\rij \gets \PhiInv( \Fhat(\xij) )$ \Comment Upper bound of z-range
    \EndFor
    \State $(\U~, \V~, \hat\sigma) \gets \arg \min
    \negthickspace\negthickspace
    \displaystyle \sum_{(i,j) \in \Omega}
    \negthickspace\negthickspace
    {-}\log \left[ \Phi\left( \frac{\rij {-} \tij}{\sigma} \right) - \Phi\left( \frac{\lij {-} \tij}{\sigma} \right) \right]$ subject to $\T=\U\V'$
    \State If necessary, adjust $(\U~,\V~)$ so that $\V~$ is orthogonal
    \Comment{See \cref{eq:orthogonalize}}
  \end{algorithmic}
\end{algorithm}

\section{XPCA Imputation}
\label{sec:mapp-copula-pred}

PCA has a large number of applications. 
In some cases, the components themselves are of interest, in which case the
factor matrices
$\U~$ and $\V~$ are used directly or as inputs to visualization, clustering, and so on.
In other cases, such as denoising 
or data imputation, we wish to provide estimates in the original data-space, and this is the
focus of this section.

Recall that PCA uses the rescaled estimate, i.e.,
$\xij~ = \hat \mu_j + \tij~  \hat v_j $ as in \cref{alg:pca-impute}. 
COCA uses $\FaltInv ( \Phi( \tij~ ) )$ as in \cref{alg:coca-impute}. 
XPCA can take a similar approach to COCA, which we call \emph{median imputation} and present in \cref{sec:median-impute}.
We also present the 
characterization of the \emph{full distribution} of $\xij*$ conditional on $\tij~$, $\Fhat_j$, and $\hat \sigma$ as described in \cref{sec:full-distribution}. 
From this information, we can infer entry means as described in
\cref{sec:mean-impute}.

\subsection{Median Imputation}
\label{sec:median-impute}

The simplest method of imputation for XPCA uses the median by simply 
evaluating 
\begin{equation}
\xij~ = \FhatInv( \Phi( \tij~ ) ). 
\end{equation}
In fact, this is the same computation as used by COCA for a given $\hat \T$..The difference is
that the loss function for the two methods is different.
We refer to this as \emph{median imputation} for the following reason:
Because $\zij*$ is normally distributed (and that distribution is symmetric),
$\tij~$ represents not only the estimated mean but also the estimated median.
Additionally, since $\Phi$ and $\FhatInv$ are both strictly increasing functions, then we can conclude that
$ \FhatInv( \Phi( \tij~ ) )$ is the estimated median of $\xij*$. 
The method is shown in \cref{alg:xpca-impute-median}.

\begin{algorithm}[t]
  \caption{XPCA Impute via Median}\label{alg:xpca-impute-median}
  Let $\mathcal{S}$ denote the entries to be imputed
  and $\U~,\V~, \hat \sigma, \Fhat_j$ be from XPCA in \cref{alg:xpca}.
  \begin{algorithmic}[1]
    \For {$(i,j) \in \mathcal{S}$} \Comment $\mathcal{S}$ denotes subset of entries to infer
    \State $\tij~ \gets \sum_{\ell=1}^k \hat u_{i\ell} \hat v_{j \ell}$ \Comment{Calculate single entry of $\T~=\U~\V~'$}
    \State $\yij~ \gets \Phi(\tij~)$ \Comment Transform standard normal to standard uniform
    \State $\xij~ \gets \FhatInv_j(\yij~)$ \Comment Transform standard uniform to original distribution
    \EndFor
  \end{algorithmic}
\end{algorithm}

\subsection{Full Distribution}
\label{sec:full-distribution}

One benefit of XPCA is that we can calculate the full conditional distribution for any entry.
By construction, the range of $\FhatInv_j$ is $\Cj$.
For every $\xi \in \Cj$, our model implies
\begin{equation}
P(\xij* = \xi | \tij~, \hat \sigma, \Fhat_j) = \Phi \left(  \frac{  r(\xi)  - \tij~}{\hat \sigma} \right) -  \Phi \left( \frac{  \ell(\xi)  - \tij~}{\hat \sigma} \right),
\end{equation}
where $r(\xi) = \PhiInv( \hat F_j (\xi) )$ and $\ell(\xi) = \PhiInv( \hat F_j(\xi-\epsilon ))$. (This is the same $\epsilon$ as in \cref{eq:epsilon}.)
Evaluating this for all $\xi \in \Cj$ produces the full 
distribution of $\xij* | \tij~, \hat \sigma, \Fhat_j$.
This is given in \cref{alg:xpca-impute-distribution}.

\begin{algorithm}[t]
  \caption{XPCA Impute Distributions}\label{alg:xpca-impute-distribution}
  Let $\mathcal{S}$ denote the entries to be imputed and
  $\U~,\V~, \hat \sigma, \Fhat_j, \epsilon$ be from XPCA in \cref{alg:xpca}.
  \begin{algorithmic}[1]
    \For {$(i,j) \in \mathcal{S}$} 
    \State{$\tij~ \gets \sum_{\ell=1}^k \hat u_{i\ell} \hat v_{j \ell}$ \Comment{Calculate single entry of $\T~=\U~\V~'$}}
    \For {$\xi \in \Cj$} \Comment{$\Cj$ is the set of distinct values in column $j$ of $\X$}
    \State{$\ell(\xi) \gets \PhiInv(\Fhat_j(\xi-\epsilon))$}
    \State{$r(\xi) \gets \PhiInv(\Fhat_j(\xi))$}
    \State{$p_{ij}(\xi) \gets \Phi \left( \frac{r(\xi) - \tij~}{\hat\sigma} \right) - \Phi \left( \frac{\ell(\xi) - \tij~}{\hat\sigma} \right)$}
    \Comment{$P(\xij* = \xi |  \tij~, \hat \sigma, \Fhat_j)$}
    \EndFor    
    \EndFor
  \end{algorithmic}
\end{algorithm}

Let $d = \max_j |\Cj| \leq m$ be the maximum number of distinct entries for any variable, and let $s=|S| \leq mn$ be the number of entries to impute. The computational cost of \cref{alg:xpca-impute-distribution} is
$O((d+k)s)$.
This may be prohibitively expensive if either $d$ or $s$ is large, so we present some alternatives in the next subsection.

\subsection{Mean Imputation}
\label{sec:mean-impute}

A natural point estimate to characterize a distribution is the mean. 
In the case of binary variables, this fully characterizes the distribution, 
while the median is simply a 0 or 1 (or $\tfrac{1}{2}$ if there is a tie). 

The full conditional distribution can be used in a straightforward way to calculate the mean:
\begin{equation}\label{eq:mean}
\mathbb{E}[ \xij* | \tij~, \hat \sigma, \Fhat_j ] = \sum_{\xi \in \Cj}  \xi \times P(\xij* = \xi |  \tij~, \hat \sigma, \Fhat_j) .
\end{equation}
This is given in \cref{alg:xpca-impute-mean}.
The cost per entry is dominated by the cost of computing the distribution, i.e., $O(d+k)$.
If we want to estimate every entry, then the work is $O((d+k)mn)$.

\begin{algorithm}[t]
  \caption{XPCA Impute via Mean}\label{alg:xpca-impute-mean}
    Let $\mathcal{S}$ denote the entries to be imputed
    and $\set{ p_{ij}}$ be from \cref{alg:xpca-impute-distribution}
  \begin{algorithmic}[1]
    \For {$(i,j) \in \mathcal{S}$} \Comment $\mathcal{S}$ denotes subset of entries to infer
    \State $\xij~ \gets \sum_{\xi \in \Cj} p_{ij}(\xi) \times \xi$
    \EndFor
  \end{algorithmic}
\end{algorithm}

If we intend to estimate all the entries in one or more columns, then we propose
an alternate strategy that may be more efficient for large $m$.
The idea is to avoid computing the full distribution for every entry in a column
but instead
compute an estimate of the expected value function with respect to $\theta$.
The expected value is a strictly increasing univariate function 
of $\theta$ that should be smooth. In fact, it can be viewed as a special type of kernel smoother, 
as it is a weighted average of the all the observed values.
We denote this function as $g_j(\theta)$ and define it to be
\begin{displaymath}
  g_j(\theta) \equiv \mathbb{E}(\RV{x} | \theta, \Fhat_j, \hat \sigma)
\end{displaymath}
To estimate $g$, we evaluate the expectation at $q$ values of $\theta$ and then use linear interpolation to fill out the rest of the function. This estimate $\hat g$ can then be used to estimate the means, i.e., the expectations for specific values of $\tij$ which is the imputation for $\xij*$.
The method for all columns is shown in \cref{alg:xpca-impute-means}. It still costs $O(mnk)$ to compute $\T~$, but we reduce the cost of the estimates from $O(dmn)$ to $O(dqn)$.
By default, we use $q = m/10$.

\begin{algorithm}[t]
  \caption{XPCA Impute Means for Entire Columns}\label{alg:xpca-impute-means}
  Let $\U~,\V~, \hat \sigma, \Fhat_j$ be from XPCA in \cref{alg:xpca}.
  \begin{algorithmic}[1]
    \State $\T~ \gets \U~\V~'$
    \For {$j = 1,\dots,n$}
    \For{$\ell = 1,\dots,q$} \Comment{$q$ is the number of interpolation points}
    \State{$\tau_{\ell} \gets \PhiInv((\ell-1)/(q-1))$}
    \State{$\varrho_{\ell} = \sum_{\xi \in \Cj} \xi \times P(\RV{x} = \xi | \tau_{\ell}, \hat \sigma, \Fhat_j)$}
    \Comment{Estimate for $\mathbb{E}(\RV{x}|\tau, \hat \sigma, \Fhat_j)$}
    \EndFor
    \For{$i=1,\dots,m$}
    \State{$\ell \gets$ index such that $\tij~ \in [\tau_\ell , \tau_{\ell+1})$}
    \State{$\alpha \gets$ value such that $\tij~ = \tau_{\ell} + \alpha (\tau_{\ell-1}-\tau_{\ell})$}
    \State{$\xij~ \gets\varrho_{\ell} + \alpha (\varrho_{\ell+1} - \varrho_{\ell})$}
    \Comment{Linearly interpolate from precomputed values}
    \EndFor
    \EndFor
  \end{algorithmic}
\end{algorithm}

\section{Examples}
\label{sec:examples}

We present the use of XPCA on both simulated datasets and real data sets. 

\subsection{Simulated Examples}
\label{sec:simulated-examples}

We compare PCA, COCA, and XPCA on simplistic simulated data using different marginal distributions. 
First, we consider the case where all
observed marginal variables are Gaussian so that all of the models have proper assumptions. 
Second, we consider exponential marginals so that PCA's assumption of Gaussian marginals
is violated.
Third, we consider the case
where 50\% of the variables are binary and 50\% are Gaussian, in which case only XPCA's assumptions are valid. 

\begin{figure}[th]
  \centering
  \subfloat[Gaussian]{
    \begin{tikzpicture}
      \begin{axis}[cycle list name=simulation,
        ytick = {0, 0.05, 0.1, 0.15},
        yticklabels={0.00, 0.05, 0.10, 0.15},
        ]
        \addplot table[x=Size,y=nDOP] {simulation.dat};
        \addplot table[x=Size,y=nDOC] {simulation.dat};
        \addplot table[x=Size,y=nDOX] {simulation.dat};
        \addplot table[x=Size,y=nDMP] {simulation.dat};
        \addplot table[x=Size,y=nDMC] {simulation.dat};
        \addplot table[x=Size,y=nDMX] {simulation.dat};
        \addplot table[x=Size,y=nMOP] {simulation.dat};
        \addplot table[x=Size,y=nMOC] {simulation.dat};
        \addplot table[x=Size,y=nMOX] {simulation.dat};
        \addplot table[x=Size,y=nMMP] {simulation.dat};
        \addplot table[x=Size,y=nMMC] {simulation.dat};
        \addplot table[x=Size,y=nMMX] {simulation.dat};
      \end{axis}
    \end{tikzpicture}
    }
  \subfloat[Exponential]{
    \begin{tikzpicture}
      \begin{axis}[cycle list name=simulation,
        ytick = {0, 0.1, 0.2, 0.3},
        yticklabels={0.00, 0.10, 0.20, 0.30},
        ]
        \addplot table[x=Size,y=eDOP] {simulation.dat};
        \addplot table[x=Size,y=eDOC] {simulation.dat};
        \addplot table[x=Size,y=eDOX] {simulation.dat};
        \addplot table[x=Size,y=eDMP] {simulation.dat};
        \addplot table[x=Size,y=eDMC] {simulation.dat};
        \addplot table[x=Size,y=eDMX] {simulation.dat};
        \addplot table[x=Size,y=eMOP] {simulation.dat};
        \addplot table[x=Size,y=eMOC] {simulation.dat};
        \addplot table[x=Size,y=eMOX] {simulation.dat};
        \addplot table[x=Size,y=eMMP] {simulation.dat};
        \addplot table[x=Size,y=eMMC] {simulation.dat};
        \addplot table[x=Size,y=eMMX] {simulation.dat};
      \end{axis}
    \end{tikzpicture}
    }
  \subfloat[50\% binary and 50\% Gaussian]{
    \begin{tikzpicture}
      \begin{axis}[cycle list name=simulation,
        legend pos=outer north east,
        ytick = {0, 0.05, 0.1, 0.15},
        yticklabels={0.00, 0.05, 0.10, 0.15},
        legend style={
          font=\tiny
        }]
        \addplot table[x=Size,y=bDOP] {simulation.dat};
        \addplot table[x=Size,y=bDOC] {simulation.dat};
        \addplot table[x=Size,y=bDOX] {simulation.dat};
        \addplot table[x=Size,y=bDMP] {simulation.dat};
        \addplot table[x=Size,y=bDMC] {simulation.dat};
        \addplot table[x=Size,y=bDMX] {simulation.dat};
        \addplot table[x=Size,y=bMOP] {simulation.dat};
        \addplot table[x=Size,y=bMOC] {simulation.dat};
        \addplot table[x=Size,y=bMOX] {simulation.dat};
        \addplot table[x=Size,y=bMMP] {simulation.dat};
        \addplot table[x=Size,y=bMMC] {simulation.dat};
        \addplot table[x=Size,y=bMMX] {simulation.dat};
        \legend{PCA, COCA, XPCA        } %
      \end{axis}
    \end{tikzpicture}
    }
  \caption{Standardized MSE for different simulated data types, using a rank-3 model with $\sigma^2=0.25$. We report the mean over 8 scenarios. We hold out 50\% of the data for validation. We compare PCA (green), COCA (blue), and XPCA (red) with respect to the observed data (circles) and underlying mean (squares) for both the data used for fitting the model (solid) and holdout validation data (dashed).}
  \label{fig:simulations}
\end{figure}
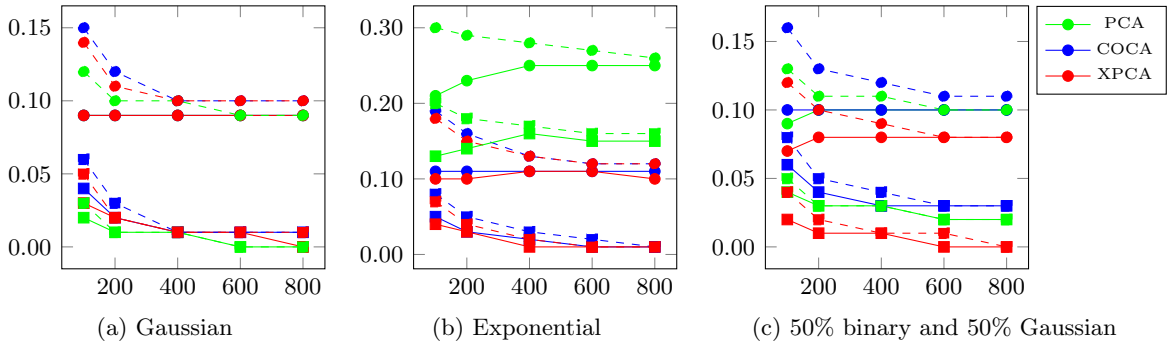

In all scenarios, data was first simulated through a rank-3 PCA model with $\sigma^2 = 0.25$ 
(i.e., 75\% of the variance in the PCA model can be explained by the low-rank structure)
and then transformed by pushing through the appropriate marginal inverse CDFs. 
We tested on square matrices 
of sizes 50, 100, 200, 400 and 800, with 50\% of the data masked completely at random. 
Each scenario was repeated 8 times. We report mean squared error (MSE) on the \emph{standardized} data, for both the observed data and the underlying means that were inputs to the observed data generation.
The results are shown in \cref{fig:simulations}.

Based on our simulations, when the data is truly from a low-rank Gaussian copula with normal marginal distributions,
PCA has the lowest MSE. This is not surprising, 
as PCA's stronger assumptions of normal marginals is indeed correct in this 
case. When the marginal distributions are exponential, both XPCA and COCA have 
significantly better MSE than PCA.
Even though PCA should yield the optimal MSE, this is under a \emph{linear} relationship between
the principal components and the data, whereas COCA and XPCA has a \emph{nonlinear} relationship thanks
to the EDF transforms.
As expected, XPCA and COCA are very similar in performance.
Finally, in the case that half the data was binary and half Gaussian,
XPCA greatly outperforms both COCA and PCA.

\pgfplotscreateplotcyclelist{simulation}{%
  green,mark=*,solid\\%
  blue,mark=*,solid\\%
  red,mark=*,solid\\%
  green,mark=*,dashed\\%
  blue,mark=*,dashed\\%
  red,mark=*,dashed\\%
  green,mark=square*,solid\\%
  blue,mark=square*,solid\\%
  red,mark=square*,solid\\%
  green,mark=square*,dashed\\%
  blue,mark=square*,dashed\\%
  red,mark=square*,dashed\\%
  }

\subsection{U.S. Senate Vote Data}
\label{sec:u.s.-senate-vote}
Our first real-world example is U.S. senate voting data from January 1989 to January 2017. This is a large span of data that covers the 101st congress
through the 114th congress.
We obtained the data from the \cite{GOV}.
It encompasses 271 senators and 9044 separate votes (the variables).
Because different senators get voted in and out of office every two years, the data is $63\%$ missing.
The entirety of this data is discrete since senators have to either vote yes ($1$), no ($-1$), or abstain ($0$).
Most previous works on analysis of senate data has only looked at a single session \cite{JaBuPiBr09,Ne14,Ja01}, so our approach adds the novelty
of multiple sessions.

We compare all the methods for different ranks, reporting standardized MSE. 
Examining in-sample errors, XPCA has a slightly lower MSE than PCA, and COCA has the highest MSE. 

Although these methods have comparable in-sample MSE, PCA has does much worse in cross-validation error. It severely overfits after rank-1 fit, going up to 15 MSE. For sake of completeness,
we do include its error in our plot but note that it is even worse than just comparing to the mean. With XPCA and COCA, 
we see a much more stable imputation at these higher ranks. One could potentially argue that a rank-1 fit is all that is needed, in which case PCA would suffice. However, senators' voting decisions
are representative of a complex decision making process; although
senators have a political party that they tend to vote with, they have a richer dimensionality to their decisions which both
XPCA and COCA show.

  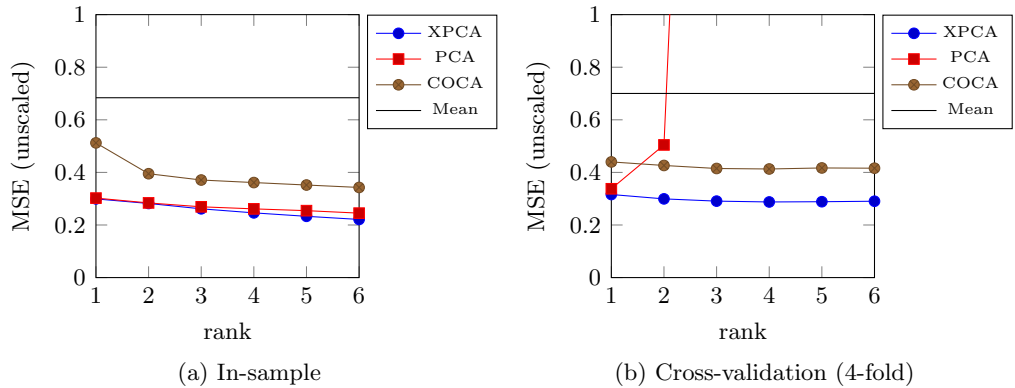
\begin{figure}[ht]
    \centering
    \pgfplotsset{
      ymode=normal,
      ymax=1,
      ymin=0,
      clip,
      restrict y to domain*=0:5,
      legend style={
        font=\tiny
      },
      xmin = 1,
      xmax = 6,
      xtick={1,2,...,6},
      legend pos=outer north east,
    }
    \subfloat[In-sample]{
      \begin{tikzpicture}
        \begin{axis} [xlabel={rank}, ylabel={MSE (unscaled)}]
          \pgfplotstableread[col sep=comma]{senator_insample_mse_unscaled.csv}\datatable;
          \pgfplotstabletranspose[colnames from=colnames]\otherdatatable{\datatable};
          \addplot table[x=colnames,y=xpca]{\otherdatatable};          
          \addplot table[x=colnames,y=pca]{\otherdatatable};          
          \addplot table[x=colnames,y=coca]{\otherdatatable};          
          \addplot[no marks] table[x=colnames,y=colmean]{\otherdatatable};
          \legend {XPCA,PCA,COCA,Mean}
        \end{axis}
      \end{tikzpicture}
      }
   \subfloat[Cross-validation (4-fold)]{
      \begin{tikzpicture}
        \begin{axis}[clip,xlabel={rank}, ylabel={MSE (unscaled)}]
          \pgfplotstableread[col sep=comma]{senator_cv_mse_unscaled.csv}\datatable;
          \pgfplotstabletranspose[colnames from=colnames]\otherdatatable{\datatable};
          \addplot table[x=colnames,y=xpca]{\otherdatatable};          
          \addplot table[x=colnames,y=pca]{\otherdatatable};          
          \addplot table[x=colnames,y=coca]{\otherdatatable};          
          \addplot[no marks] table[x=colnames,y=colmean]{\otherdatatable};          
          \legend {XPCA,PCA,COCA,Mean}
        \end{axis}
      \end{tikzpicture}
      }
    \caption{MSE (unscaled) on Senator Vote Data}
    \label{fig:senator-mse}
  \end{figure}

\begin{figure}[tp]
  \centering
  \newcommand{\DemVote}[1]{\textcolor{blue}{#1 $\geq$ 50\%}}
  \newcommand{\RepVote}[1]{\textcolor{red}{#1 $\geq$ 50\%}}
  \newcommand{\VoteCombo}[2]{\textcolor{blue}{#1$\geq$50\%},\textcolor{red}{#2$\geq$50\%}}
\includegraphics[width=\textwidth]%
{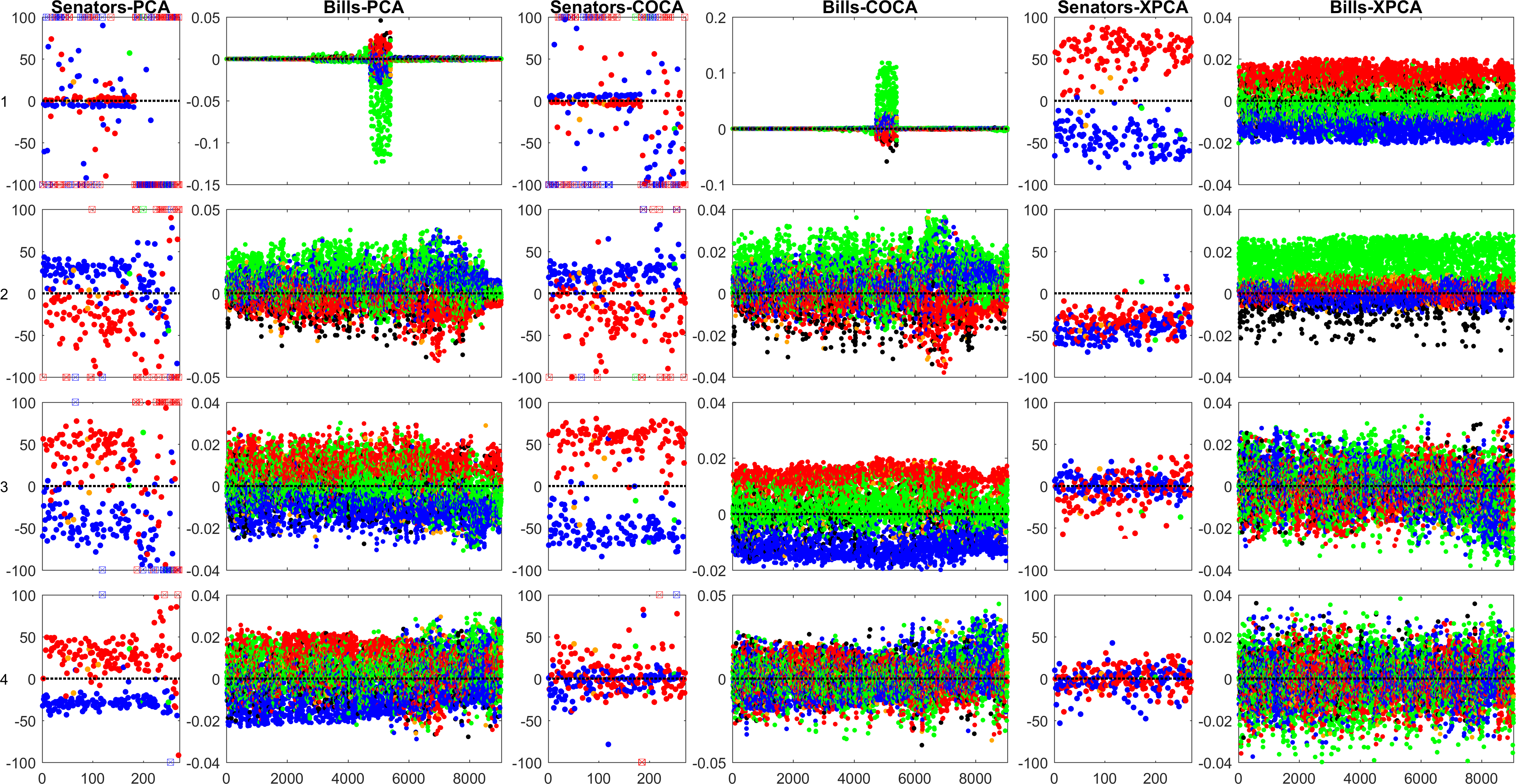}
\begin{tikzpicture}[node distance=3mm, inner sep=0pt, minimum size=1mm,font=\tiny,
  symbol/.style={circle,draw=#1,fill=#1,node distance=1mm},
  senator/.style={circle,draw=#1,fill=#1},
  vote/.style={circle,draw=#1,fill=#1},
  label/.style={node distance=5mm},
  ]
  \draw[white,very thin] (0,0) grid (10,1);
  \node[left] at (1,0.5) (sntr-label) {Senators:};
  \node[label] (sntr-blue) [right=of sntr-label] {Democrat}; 
  \node[symbol=blue] [left=of sntr-blue] {};
  \node[label] (sntr-red) [right=of sntr-blue] {Republican}; 
  \node[symbol=red] [left=of sntr-red] {};
  \node[label] (sntr-green) [right=of sntr-red] {Independent}; 
  \node[symbol=green] [left=of sntr-green] {};
  \node[label] (sntr-orange) [right=of sntr-green] {Switched Parties}; 
  \node[symbol=orange] [left=of sntr-orange] {};
  \node[label] (sntr-outside) [right=of sntr-orange] {Outside Displayed Range};
  \node[node distance=1mm,scale=0.8] [left=of sntr-outside] {$\boxtimes$};
  \node[left] at (1,0.1) (vote-label) {Votes:};
  \node[label] (vote-green) [right=of vote-label] {\VoteCombo{Yes}{Yes}};
  \node[symbol=green] [left=of vote-green] {};
  \node[label] (vote-blue) [right=of vote-green] {\VoteCombo{Yes}{No}};
  \node[symbol=blue] [left=of vote-blue] {};
  \node[label] (vote-red) [right=of vote-blue] {\VoteCombo{No}{Yes}};
  \node[symbol=red] [left=of vote-red] {};
  \node[label] (vote-black) [right=of vote-red] {\VoteCombo{No}{No}};
  \node[symbol=black] [left=of vote-black] {};
  \node[label] (vote-orange) [right=of vote-black] {None of the Above};
  \node[symbol=orange] [left=of vote-orange] {};
\end{tikzpicture}
\caption{Components of a rank 4 XPCA decomposition according to Algorithm \cref{alg:xpca-impute-mean}. On the left hand side are the components of the 9044 bills and the right hand side are the 271 senators. XPCA shows separation of republican and democrats in both senator political parties and republican, democrat, bipartisan supported bills. The colorings for the senators is red indicates she voted only as a republican, blue indicates she voted only as a democrat, green indicates sshe voted as only as an independent, and orange indicates inconsistent party membership.  The colorings for the bills are red indicates that $>50\%$ of Democrats voted yes and $>50\%$ of Republicans voted no, blue indicates $>50\%$ of Republicans voted yes and $>50\%$ of Democrats voted no,  green indicates both $>50\%$ of Democrats voted yes and $>50\%$ of Republicans voted yes, black indicates both $>50\%$ of Democrats voted no and $>50\%$ of Republicans vote no, and yellow indicates the bill had no majority split.}
\label{fig:xpca_components}
\end{figure}

Although XPCA's mean squared error is lower than COCA, the real advantage is seen when looking deeper at the in-sample decompositions.
In \cref{fig:xpca_components}, we show the components from a rank 4 decomposition of PCA using \cref{alg:pca-impute},  COCA using \cref{alg:coca-impute}, and XPCA using \cref{alg:xpca-impute-mean}. Visualization of the XPCA components shows the clear separation of political party affiliation as well as bipartisan supporters. As shown in the first component in the first row of plots, XPCA demonstrates separation of senators by their political party as well as their corresponding party-dictated voting behavior. In the second component of XPCA in the second row of plots, there is a clear distinction of senators bipartisan voting behavior. In addition, there is a visible shift between red and blue approximately at bill index 2000 and again at 6000 and 9000. This represents the party control over the senate: if republicans had majority vote, the red points get larger values than blue. These are fundamental voting patterns not as clearly shown in PCA nor COCA. 
In both COCA's and PCA's components, there is a separation of republican vs democrats in the third component, and arguably a bipartisan voting behavior represented in the second component. However, both algorithms demonstrate a spike in value ranges for the first component contained to the 108th congress. We hypothesize that this would be fixed by regularization, but the cause of both algorithms demonstrating this behavior is unknown. The 108th congress was removed from the data and PCA no longer had this anomalous range spike, perhaps indicating that the 108th congress is unique in PCA's analysis. However, when the 108th congress was removed and COCA analyzed this special view of the data, it moved the range spike to a different congress block. This suggests that COCA isn't picking up on unique congressional sessions, but actually a victim of lack of regularization.

\subsection{Basketball Data Set}
\label{sec:basketball-data-set}
Our next set of experiments involves %
statistics on NBA players from the 2015-2016 season. 
This dataset comprises 40 variables on 476 players.
The variables include performance summaries over the entire season such as
shots scored, number of offensive and defensive rebounds, and number of assists.
Additionally, 
a few out-of-game statistics are included, such as height, weight, draft round, and draft number. 
Two binary variables were included: most valuable player (MVP) and defensive MVP. 
These were binary variables that indicated if the player had \emph{ever} been voted
MVP or defensive MVP in any season, up to the 2015-2016 season. 
The dataset is complete, i.e., no missing data and was compiled from two sources, namely \cite{NBA} and \cite{MVP, DPA}.

\subsubsection{Data Imputation for Basketball Data Set}
\label{sec:data-imputation}

We examine imputation error by
performing 20-fold cross validation (i.e., randomly divide the data into 20 parts and hold out on part for imputation in each test) to compute the imputation error.
We compute each factorization for ranks 1 through 10
and then impute the held out entries (using mean imputation for XPCA per \cref{alg:xpca-impute-means}).
As in our previous experiments, we compare the standardized MSE.
The results can be seen in \cref{table:BB_RMSE}.

 \begin{table}[ht]
 \centering
 \begin{tabular}{rrrrr}
   \hline
  & XPCA & PCA & COCA & Column Mean \\ 
   \hline
 Rank 1 &  \textbf{0.466}  & 0.480 & 0.489 & 1.005 \\ 
   Rank 2 & \textbf{0.350} & 0.379 & 0.364 & 1.005 \\ 
   Rank 3 & \textbf{0.330} & 0.365 & \textbf{0.330}& 1.005 \\ 
   Rank 4 & \textbf{0.319} & 0.341 & 0.377 & 1.005 \\ 
   Rank 5 &\textbf{0.316} & 1.476 & 0.383 & 1.005 \\ 
   Rank 6 & \textbf{0.340} & 3.307 & 0.430 & 1.005 \\ 
   Rank 7 & \textbf{0.344} & 11.754 & 0.532 & 1.005 \\ 
   Rank 8 & \textbf{0.329} & 14.030 & 0.653 & 1.005 \\ 
   Rank 9 & \textbf{0.353} & 9.287 & 0.717 & 1.005 \\ 
   Rank 10 & \textbf{0.336} & 53.885 & 0.717 & 1.005 \\ 
    \hline
 \end{tabular}
 \caption{Rescaled MSE by decomposition type} 
 \label{table:BB_RMSE}
 \end{table} 
 
The rank-5 XPCA decomposition is the best overall MSE, with 4\% improvement 
versus the best COCA fit and 8\% improvement versus the best PCA fit. 
All three methods showed evidence
of over-fitting by rank 6. 
However, PCA has more extreme problems with overfitting
than either COCA or XPCA. This is
most likely due to the fact that the XPCA and COCA models only impute 
values within the observed range for each column of data 
while nothing bounds the estimates produced by PCA. 
Interestingly, the effect of overfitting still seems much lower for XPCA than COCA,
even though both only impute bounded estimates. 

We examine some of individual estimates in more detail, showing the imputation results in the original dataspace.
Of the 476 players in 2015-2016 season, there were six MVPs and five defensive MVPs.
\Cref{fig:imputedPointEstimates}
shows a few examples of imputed values produced when
these values were masked in the cross validation procedure.
We used the rank that corresponded with best cross-validation error for a given method: rank 5 for XPCA, 
rank 4 for PCA, and rank 3 for COCA.
We show results
for three players: Kevin Durant, 
a former MVP; Dwight Howard, a former defensive MVP; and Anthony Morrow, a more typical NBA player. 

\begin{table}[ht]
\small
\centering
 \begin{tabular}{ccc | cc| cc}
   \hline
   & \multicolumn{2}{c}{Kevin Durant} & \multicolumn{2}{c}{Dwight Howard}  & \multicolumn{2}{c}{Anthony Morrow} \\
   & MVP & Defensive MVP & MVP & Defensive MVP & MVP & Defensive MVP  \\ 
   \hline True Values & 1 & 0 &  0 & 1 & 0 & 0\\ 
   XPCA Estimates & \bf 0.383 & 0.048 & \bf 0.000 & \bf 0.204  & \bf 0.000 & \bf 0.000  \\ 
   PCA Estimates & 0.142 & 0.091 & 0.142 & 0.048 & -0.087 & -0.057  \\ 
   COCA Estimates & 0 & \bf 0 & \bf 0 & 0 & \bf 0 & \bf 0 \\ 
    \hline
 \end{tabular}
 \caption{Imputed MVP Variables for Kevin Durant,  Dwight Howard and Anthony Morrow} 
 \label{fig:imputedPointEstimates}
 \end{table}

Per \cref{alg:xpca-impute-mean}, XPCA can only impute
values within the range of possible values, and these can be interpreted as proper probabilities in this case of
binary variables.
In contrast, PCA's estimate from \cref{alg:pca-impute} cannot be interpreted as probabilities, as evidenced by the negative 
values for Anthony Morrow.
COCA can only estimate observed values per \cref{alg:xpca-impute-means}, and in this case all four estimated values are zero. In fact, 
COCA imputed zeros for both variables \emph{for all players}.

\subsubsection{Components for Basketball Data Set}
\label{sec:components}

We examine the components from the rank-three decompositions and present the top ten
variables (in absolute value) for each component in \cref{tab:top_facts}.
The full components for each decomposition can be found in \cref{sec:full-comp-bask}. 

For all three methods, the first component is strongly related to the minutes played which in turn
correlates with count variables (as opposed to percentages). The second factor 
appears strongly related with the size of the player (i.e., height and weight), which appears to favor 
defensive plays (blocks, rebounds) but less favorable for more offensive statistics (three point shots). 
For the third factor, the methods appear to capture the value of a player, although this is a little less straightforward for PCA. 
The third factor for PCA does not have a clean interpretation.
The third factor for COCA and XPCA seem to relate to the player value. We see that the draft number and round are the most important features, and that these are inversely related to the MVP status (XPCA also picks up DefensiveMVP). This relationship is sensible since the strongest players will generally have had the lowest draft numbers.

\begin{table}[ht]
\tiny
\centering
\begin{tabular}{rlrlrlr}
  \hline
\multicolumn{7}{c}{PCA}\\
  \hline
 & PC 1:Name & PC 1:Value & PC 2:Name & PC 2:Value & PC 3:Name & PC 3:Value \\ 
  \hline
  1 & MinutesPlayed & 0.208 & Weight & 0.351 & TripleDouble & 0.402 \\ 
  2 & OpponentPtsDuringPlay & 0.207 & OffensiveRebound & 0.319 & GamesPlayed & -0.329 \\ 
  3 & ShotsMade & 0.207 & Height & 0.317 & TeamLosses & -0.324 \\ 
  4 & Points & 0.207 & Block & 0.284 & DoubleDouble & 0.286 \\ 
  5 & Opponent2ndPts & 0.205 & ThreePTsAttempted & -0.279 & Assist & 0.222 \\ 
  6 & OpponentsTOPoints & 0.205 & ThreePTsMade & -0.275 & FreeThrowsMade & 0.218 \\ 
  7 & ShotsAttempted & 0.205 & ThreePTsPercentage & -0.242 & FreeThrowsAttempted & 0.218 \\ 
  8 & OpponentFastBreakPts & 0.202 & TotalRebound & 0.231 & PersonalFouls & -0.205 \\ 
  9 & PointsTurnOver & 0.198 & ShotsPercentage & 0.228 & TimesFouled & 0.188 \\ 
  10 & Turnover & 0.196 & DoubleDouble & 0.210 & TeamWins & -0.187 \\    \hline

\multicolumn{7}{c}{COCA}\\

  \hline
 & PC 1:Name & PC 1:Value & PC 2:Name & PC 2:Value & PC 3:Name & PC 3:Value \\ 
  \hline
1 & Points & 0.199 & Height & 0.384 & draftNumber & 0.601 \\ 
  2 & ShotsMade & 0.199 & Weight & 0.374 & draftRound & 0.588 \\ 
  3 & MinutesPlayed & 0.199 & ThreePTsMade & -0.327 & MVP & -0.280 \\ 
  4 & OpponentPtsDuringPlay & 0.198 & ThreePTsAttempted & -0.324 & TeamLosses & 0.232 \\ 
  5 & ShotsAttempted & 0.198 & ShotsPercentage & 0.277 & TripleDouble & -0.199 \\ 
  6 & OpponentsTOPoints & 0.197 & ThreePTsPercentage & -0.257 & GamesPlayed & 0.170 \\ 
  7 & Opponent2ndPts & 0.196 & OffensiveRebound & 0.246 & PersonalFouls & 0.125 \\ 
  8 & PointsTurnOver & 0.194 & Block & 0.231 & FreeThrowPercentage & -0.112 \\ 
  9 & TimesFouled & 0.194 & FreeThrowPercentage & -0.200 & DoubleDouble & -0.102 \\ 
  10 & OpponentFastBreakPts & 0.194 & DoubleDouble & 0.185 & OffensiveRebound & 0.091 \\ 
     \hline

\multicolumn{7}{c}{XPCA}\\

  \hline
 & PC 1:Name & PC 1:Value & PC 2:Name & PC 2:Value & PC 3:Name & PC 3:Value \\ 
  \hline
1 & TripleDouble & 0.198 & Height & 0.378 & draftRound & 0.543 \\ 
  2 & Points & 0.194 & Weight & 0.369 & draftNumber & 0.476 \\ 
  3 & ShotsMade & 0.194 & DefensiveMVP & 0.331 & MVP & -0.450 \\ 
  4 & MinutesPlayed & 0.193 & ThreePTsMade & -0.303 & DefensiveMVP & -0.247 \\ 
  5 & ShotsAttempted & 0.192 & ThreePTsAttempted & -0.299 & TeamLosses & 0.207 \\ 
  6 & OpponentPtsDuringPlay & 0.192 & ShotsPercentage & 0.264 & FreeThrowPercentage & -0.200 \\ 
  7 & OpponentsTOPoints & 0.190 & OffensiveRebound & 0.221 & GamesPlayed & 0.129 \\ 
  8 & PointsTurnOver & 0.190 & ThreePTsPercentage & -0.210 & PersonalFouls & 0.126 \\ 
  9 & TimesFouled & 0.190 & Block & 0.210 & OffensiveRebound & 0.125 \\ 
  10 & Opponent2ndPts & 0.189 & FreeThrowPercentage & -0.170 & ShotsPercentage & 0.104 \\    \hline
\end{tabular}
\caption{Top ten factors per decomposition method} 
\label{tab:top_facts}
\end{table}

Although in general it is difficult to argue that one set of components is definitively \emph{better} than another,
we contend XPCA does better at finding relations in situations where there is a \emph{heavy atom},
i.e., a particular value with very high probability, e.g., 75\% or higher.

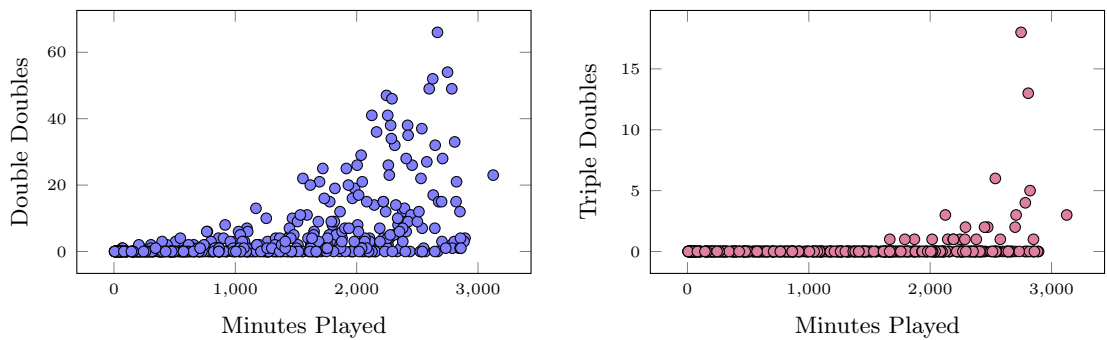
\begin{figure}
  \centering
  \begin{tikzpicture}
    \begin{axis}[
      scatter/use mapped color={draw=black,fill=blue!50},
      xlabel={Minutes Played},
      ylabel={Double Doubles},
      tick label style={font=\tiny},
      label style={font=\footnotesize},
      width=3in,
      ]
      \pgfplotstableread[col sep=comma]{nba_data_renamed.csv}\datatable;
      \addplot[ scatter, only marks, mark size = 2pt ]
      table[x="MinutesPlayed",y="DoubleDouble"]{\datatable};
    \end{axis}
  \end{tikzpicture}
~~
  \begin{tikzpicture}
    \begin{axis}[
      scatter/use mapped color={draw=black,fill=purple!50},
      xlabel={Minutes Played},
      ylabel={Triple Doubles},
      tick label style={font=\tiny},
      label style={font=\footnotesize},
      width=3in,
      ]
      \pgfplotstableread[col sep=comma]{nba_data_renamed.csv}\datatable;
      \addplot[ scatter, only marks, mark size = 2pt ]
      table[x="MinutesPlayed",y="TripleDouble"]{\datatable};
    \end{axis}
  \end{tikzpicture}
  \caption{From the basketball data set, double and triple doubles
    versus minutes played. Clearly the number of doubles is correlated
    with minutes played, but the triple doubles are rare.}
  \label{fig:TDandDD}
\end{figure}

Consider the influence of double and triple doubles in the first component. In basketball, 
a double double is when a player gets double digit statistics in two positive categories during a single game, and  
a triple double is when a player gets double digit statistics in three positive categories. 
Hence, the double doubles must be less than or equal to the triple doubles.
In our dataset, 232 of 476 players had at least one double double, while only 12 had at least one triple double. 
Recall that the first component was primarily about minutes played and related count variables.
All three methods had moderate to heavy loadings for Double Double in this component
(PCA = 0.137, COCA = 0.143 and XPCA = 0.164), %
but only XPCA put heavy loading on Triple Doubles 
(PCA = 0.067, COCA =  0.071 and XPCA = 0.198). 
In \cref{fig:TDandDD}, we show correlation between minutes played and either double or triple doubles.
We hypothesize that triple doubles have a heavy atom at zero, which makes it much more difficult
for PCA or COCA to find the correlation between triple doubles and minutes played.

As another example, consider the defensive MVP variable in the second component.
Recall that this factor seems to split defensive and offensive players, giving positive weights
for defensive traits (rebounds, blocks) and negative weights for offensive traits (three pointers, free throws).
Only XPCA puts heavy positive loading onto Defensive MVP (PCA = 0.077, COCA = 0.078, XPCA = 0.331), a binary 
variable that should be correlated to defense-related variables and for `false' is a heavy atom.

Finally, we consider the third factors for COCA and XPCA which correspond to player value in some sense since the most important characteristics are draft numbers. COCA picks up MVP (PCA = 0.173, COCA = 0.280, XPCA = 0.450) in its top ten, but XPCA gets both MVP and defensive MVP (PCA = 0.057, COCA = 0.091, XPCA = 0.247).

These examples suggest that XPCA may do a better job at finding
relations between variables with a heavy atom and other variables in
the dataset.  The reason for the difference is that for a given $\xi$
in the set of observed values for a given column, $r(\xi)$ and
$\ell(\xi)$ will be far apart. This means that the COCA and XPCA loss
functions will be markedly different.

\subsubsection{Conditional Distributions}
\label{sec:cond-distr}

\begin{figure}
  \centering
  \subfloat[Total points for all players, compiled from raw data]{\label{fig:margShots}%
    \begin{tikzpicture}
      \begin{axis}[ybar,ymin=0,width=3in,axis y line=none, axis x line=bottom]
        \pgfplotstableread[col sep=comma]{nba_data_renamed.csv}\datatable;
        \addplot+[ hist={bins=25,data min=0,data max=2300} ]
        table[y="Points"]{\datatable};
      \end{axis}
    \end{tikzpicture}}
  ~~~
  \subfloat[Estimated conditional distributions of total points for specific players]{\label{fig:condShots}%
    \begin{tikzpicture}
      \begin{axis}[ybar,bar shift=0pt,ymin=0,xmin=0,xmax=2400,width=3in,legend style={font=\tiny},legend pos=outer north east,
        axis y line=none, axis x line=bottom,legend style={at={(0.5,1.15)},
          anchor=north,legend columns=-1}]
        \pgfplotstableread[col sep=comma]{player_pdfs.dat}\datatable;
        \addplot+[bar width=7pt,opacity=0.8] table[x index=0,y index=1]{\datatable};
        \addplot+[bar width=7pt,opacity=0.8] table[x index=0,y index=2]{\datatable};
        \addplot+[bar width=7pt,opacity=0.8] table[x index=0,y index=3]{\datatable};
        \legend{Kevin Durant,Dwight Howard,Anthony Morrow}
      \end{axis}
    \end{tikzpicture}}
  \caption{Distributions of shots}
\end{figure}
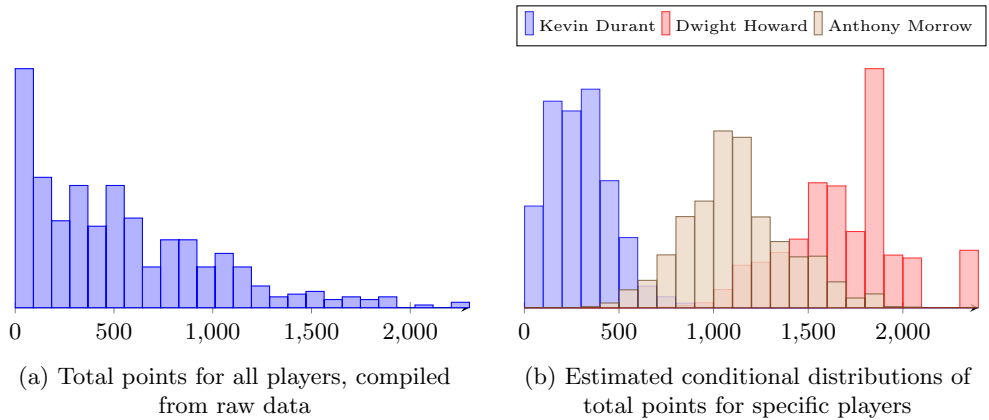

Conditional on our fitted model, we can provide an estimated distribution for a given entry. 
For reference, we first look at the marginal distribution of points, across all players, 
in \cref{fig:margShots}, in bins of size 100.
Using \cref{alg:xpca-impute-distribution}, we compute estimated conditional distributions for Kevin Durant, Dwight Howard and Anthony Morrow.
These are shown in \cref{fig:condShots}, in bins of size 100.
We caution that these distributions reflect the residual uncertainty conditional on the fitted parameters, 
rather than also incorporating the uncertainty in the parameters themselves. 

\section{Discussion}
\label{sec:discussion}
We consider the problem of heterogeneous variables in PCA analysis of data matrices whose rows correspond to objects or persons
and whose columns correspond to features or variables.
We propose XPCA, a Gaussian-copula-based decomposition method
specifically targeted at datasets with a mixture of continuous and discrete variables.
We develop this in a probabilistic framework and show that the XPCA factorization is a maximum likelihood solution.
We provide algorithms for computing XPCA and for imputing estimates of the entries in the original data matrix.
In simulations, 
we show that this enables us to relax the marginal assumptions built into both PCA and COCA so that discrete data can be incorporated.
Additionally, we show that XPCA has advantages as compared to PCA and COCA in analysis of real data sets where some variables have heavy atoms (a specific outcome with very high probability). 
We have stressed the utility of XPCA for discrete variables, but XPCA is also appropriate for semi-continuous data such as zero-inflated distributions which have a heavy atom at zero.

In future work, there are potential improvements to consider.
Decreasing the computation time is a priority since the computational cost of XPCA
can be an order of magnitude slower than a PCA or COCA decomposition using alternating least squares, but 
we hypothesize that an improved optimization algorithm would alleviate most of this problem.
In terms of imputation of missing data, we can potentially improve out-of-sample error 
by including penalization.
We also want to investigate letting $\sigma$ (the residual error remaining after the low-rank decomposition) vary
across columns, which would be closer to a factor analysis model.
Although we can handle discrete data, we have no easy way yet of incorporating categorical data.

\acks{%
  We thank David Hong for suggesting the  U.S.~Senator voting data and providing us with scripts to collect it,
  and Justin Jacobs for recommending the NBA player data and providing us with an initial version of the dataset.
  We want to acknowledge Jessica Gronski for work on a related project that helped to motivate this work.

Sandia National Laboratories is a multimission laboratory managed and operated by National Technology and Engineering Solutions of Sandia, LLC., a wholly owned subsidiary of Honeywell International, Inc., for the U.S. Department of Energy's National Nuclear Security Administration under contract DE-NA-0003525.
}

\appendix

\section{Derivatives}
\label{sec:derivatives}

Recall that $\negloglik$ is the loss function (i.e., negative log-likelihood) as defined in \cref{eq:nll-xpca} and that $\T = \U\V'$.
Define the following substitutions:
\begin{align*}
  \diffr* &= \rij - \tij, &
  \diffl* &= \lij - \tij, & 
                            \tau &= \sigma^{-1}, \\
  \rij* &= \tau \diffr*, &
                          \lij* &= \tau \diffl*, & 
  \prob* &= \Phi(\rij*) - \Phi(\lij*). &
\end{align*}
Then $\negloglik = - \sum_{(i,j) \in \Omega} \log \prob*$.
We need to compute derivatives of this with respect to $\sigma$, $\U$, and $\V$.
Recall that $\Phi'(x) = \phi(x)$ and $\phi'(x) = -x \phi(x)$.

\subsection{Derivatives with respect to $\sigma$}
We compute the derivative with respect to $\tau = 1/\sigma$ and then use the chain rule to get the derivatives with respect to $\sigma$.
The first partial derivative is
\begin{align*}
  \FD{\negloglik}{\sigma} & = - \sum_{(i,j) \in \Omega} \frac{1}{\prob*} \cdot \FD{\prob*}{\tau} \cdot \FD{\tau}{\sigma} \\
  & = - \sum_{(i,j) \in \Omega} \prob*^{-1} \cdot
    \Bigl[
      \diffr* \, \phi(\tau \diffr*) - \diffl* \, \phi( \tau \diffl*) %
    \Bigr]
    \cdot \Bigl[ -\sigma^{-2} \Bigr] \\
  & = \frac{1}{\sigma^2} \sum_{(i,j) \in \Omega}
                            \frac{1}{\prob*} \cdot
    \Bigl[
      \diffr* \, \phi(\rij*) - \diffl* \, \phi( \lij* ) %
    \Bigr].
\end{align*}
The second partial derivative w.r.t.\@ $\tau$ is
\begin{align*}
  \SD{\negloglik}{\tau} &= \sum_{(i,j) \in \Omega} \prob*^{-2} \cdot \left( \FD{\prob*}{\tau} \right)^2 - \prob*^{-1} \cdot \SD{\prob*}{\tau}\\
               &=\sum_{(i,j) \in \Omega} \prob*^{-2} \Bigl[ \diffr* \, \phi(\tau \diffr*) - \diffl* \, \phi( \tau \diffl*) \Bigr]^2
                 + \prob*^{-1} \tau \Bigl[ \diffr*^3 \, \phi(\tau\diffr*) - \diffl*^3 \, \phi(\tau\diffl*) \Bigr] .                
\end{align*}
From this, we can compute the derivatives w.r.t.\@ $\sigma$:
\begin{align*}
  \SD{\negloglik}{\sigma}
  & = \SD{\negloglik}{\tau} \cdot \left( \FD{\tau}{\sigma} \right)^2
  + \FD{\negloglik}{\tau} \cdot \SD{\tau}{\sigma} \\
  & = \SD{\negloglik}{\tau} \cdot \frac{1}{\sigma^4} 
    + \FD{\negloglik}{\tau} \cdot \frac{2}{\sigma^3} \\
  & = \frac{1}{\sigma^3}\sum_{(i,j) \in \Omega} \frac{\prob*^{-2}}{\sigma} \Bigl[ \diffr* \, \phi( \rij*) - \diffl* \, \phi( \lij* ) \Bigr]^2
    + \frac{\prob*^{-1}}{\sigma^2} \Bigl[ \diffr*^3 \, \phi(\rij*) - \diffl*^3 \, \phi(\lij*) \Bigr] \\
  & \phantom{= \frac{1}{\sigma^3} \sum_{(i,j) \in \Omega}}
    - 2 \prob*^{-1}
    \Bigl[
      \diffr* \, \phi(\rij*) - \diffl* \, \phi( \lij*) %
    \Bigr].
\end{align*}

\subsection{Derivatives with respect to $\U$ and $\V$}
If $\loss* = - \log \prob*$ is the $(i,j)$th summand, then 
${\partial \loss*} / {\partial \theta_{k\ell}}$ is zero unless $i=k$ and $j=\ell$; thus, we only consider ${\partial \loss*} / {\partial \tij}$.
The first derivative is
\begin{displaymath}
  \FD{\loss*}{\tij}
  = - \frac{1}{\prob*} \cdot \FD{\prob*}{\tij} 
  = -\frac{1}{\prob*} \cdot - \frac{1}{\sigma}\bigl[ \phi(\rij*) - \phi(\lij*) \bigr]
  = \frac{1}{\sigma\prob*} \bigl[ \phi(\rij*) - \phi(\lij*) \bigr].
\end{displaymath}
The second derivative is
\begin{align*}
  \SD{\loss*}{\tij}
  &= \prob*^{-2} \cdot \left( \FD{\prob*}{\tij} \right)^2
    -\prob*^{-1} \cdot \SD{\prob*}{\tij}\\
  &= \frac{1}{\sigma^2 \prob*^{2}} \Bigl[ \phi(\rij*) - \phi(\lij*) \Bigr]^2
    + \frac{1}{\sigma^3 \prob*} \Bigl[ \diffr* \phi(\rij*) - \diffl* \phi(\lij*) \Bigr].
\end{align*}

Now, we return to the NLL.
For a given single element $u_{i\ell}$, we have
\begin{displaymath}
  \FD{\text{NLL}}{u_{i\ell}} = \sum_{j:(i,j) \in \Omega} \FD{\loss*}{\tij} \cdot \FD{\tij}{u_{i\ell}}.
\end{displaymath}
Let $\LA$ be the $m \times n$ \emph{first-derivative} matrix whose $(i,j)$ entry is $\FD{\loss*}{\tij}$.
Then, in matrix or vector form, this looks like
\begin{displaymath}
  \FD{\text{NLL}}{\U} = \LA\V,
  \qtext{or}
  \FD{\text{NLL}}{\U_{i:}} = \LA_{i:}\V.
\end{displaymath}
These are, in turn, an $n \times k$ matrix or a $k$-vector. Note that the second version is a $1 \times n$ vector times an $n \times k$ matrix, so it results in a $1 \times k$ \emph{row} vector, which makes sense for updating a row of $\U$.
Using analogous reasoning, for $\V$ we have
\begin{displaymath}
  \FD{\text{NLL}}{\V} = \LA^T\U,
  \qtext{or}
  \FD{\text{NLL}}{\V_{j:}} = \left(\LA_{:j}\right)^T \U.
\end{displaymath}

For the second derivative, we have
\begin{displaymath}
  \SD{\text{NLL}}{u_{i\ell}}
  = \sum_{j:(i,j) \in \Omega}
  \SD{\loss*}{\tij} \cdot \left( \FD{\tij}{u_{i\ell}} \right)^2
  + \FD{\loss*}{\tij} \cdot \SD{\tij}{u_{i\ell}} .
\end{displaymath}
The second term is zero, so we look only at the first term.
We can write this for a row of $\U$ as
\begin{displaymath}
\SD{\text{NLL}}{\U_{i:}}  = \V^{T} \LB \V,
\end{displaymath}
where $\LB$ is an $n \times n$ diagonal matrix whose $j$th diagonal entry is
\begin{displaymath}
  \LB(j,j) = \SD{\loss*}{\tij}.
\end{displaymath}
In other words, $\LB$ is the diagonal created by the $i$th \emph{row} of the matrix of second derivatives w.r.t. $\tij$.
An analogous argument w.r.t.\@ $\V$ leads to
\begin{displaymath}
\SD{\text{NLL}}{\V_{j:}}  = \U^{T} \LC \U,  
\end{displaymath}
In this case, $\LC$ is the diagonal created by the $j$th \emph{column} of the matrix of second derivatives w.r.t. $\tij$.

\section{Midpoint imputation for COCA}
\label{sec:midp-imput-coca}

\citet{HaLi12} developed COCA in the context of continuous data and so do not discuss the issue of ties explicitly.
A standard approach is just to use the maximum rank for ties; however, this causes problems for COCA in the case
of discrete variables.
Instead, we suggest using the midpoint rank. 

To demonstrate that this is a good idea, we simulated an even mix of
Gaussian and binary variables in the same manner as presented in
section \cref{sec:simulated-examples} for $m=n=100$.
We perform 20-fold cross validation and compare the MSE of the two methods on the held out data in \cref{tab:tieMethods}.
Using the midpoint rank rather than the maximum results in  an  38\% reduction in the MSE.

\begin{table}[ht]
\centering
\begin{tabular}{rrr}
  \hline
 Tie Method & MSE & SE(MSE) \\ 
  \hline
Midpoint & 0.2001 & 0.0004 \\ 
Last & 0.3254 & 0.0030 \\ 
   \hline
\end{tabular}
\caption{Comparing COCA methods for ties} 
\label{tab:tieMethods}
\end{table}

Therefore, in all uses of COCA in this paper, ties are handled using the midpoint rank rather than the maximum rank.

\section{Full Components for Basket Ball Data}
\label{sec:full-comp-bask}

In \cref{sec:components}, we present only the top ten variable from each component. In this appendix, we show the complete data
in \cref{tab:bbfacts}.

\begin{table}[ht]
  \tiny
  \setlength{\tabcolsep}{4pt}
  \centering
  \begin{tabular}{r|*{3}{r@{\,}>{(}l<{)}}|*{3}{r@{\,}>{(}l<{)}}|*{3}{r@{\,}>{(}l<{)}}}
    \hline
    Variable & \multicolumn{2}{c}{PCA-1} & \multicolumn{2}{c}{PCA-2} & \multicolumn{2}{c}{PCA-3} & \multicolumn{2}{c}{COCA-1} & \multicolumn{2}{c}{COCA-2} & \multicolumn{2}{c}{COCA-3} & \multicolumn{2}{c}{XPCA-1} & \multicolumn{2}{c}{XPCA-2} & \multicolumn{2}{c}{XPCA-3} \\ \hline  
GamesPlayed            & -.176 & 21&  .011 & 37& -.329 &  2& -.175 & 22&  .009 & 37&  .164 &  6& -.167 & 24&  .026 & 33&  .130 &  7\\ 
TeamWins               & -.137 & 25&  .010 & 38& -.184 & 10& -.142 & 27&  .005 & 39&  .003 & 40& -.138 & 28&  .002 & 38& -.021 & 31\\ 
TeamLosses             & -.132 & 29&  .007 & 40& -.326 &  3& -.142 & 28&  .010 & 36&  .229 &  4& -.136 & 29&  .036 & 31&  .214 &  5\\ 
MinutesPlayed          & -.208 &  1&  .037 & 29& -.125 & 18& -.199 &  3&  .044 & 27&  .036 & 22& -.193 &  4&  .051 & 26&  .030 & 28\\ 
ShotsMade              & -.207 &  3&  .026 & 34&  .053 & 32& -.199 &  2&  .020 & 33& -.033 & 27& -.194 &  3&  .021 & 35& -.028 & 30\\ 
ShotsAttempted         & -.205 &  7&  .076 & 20&  .034 & 35& -.198 &  5&  .064 & 22& -.038 & 20& -.192 &  5&  .063 & 22& -.040 & 22\\ 
ShotsPercentage        & -.050 & 35& -.225 &  9& -.043 & 34& -.057 & 34& -.277 &  5&  .079 & 12& -.056 & 35& -.265 &  6&  .099 & 12\\ 
ThreePTsMade           & -.133 & 28&  .271 &  6& -.088 & 24& -.119 & 30&  .327 &  3& -.024 & 31& -.117 & 31&  .303 &  4& -.101 & 10\\ 
ThreePTsAttempted      & -.138 & 24&  .275 &  5& -.089 & 23& -.123 & 29&  .324 &  4& -.036 & 23& -.119 & 30&  .299 &  5& -.101 & 11\\ 
ThreePTsPercentage     & -.048 & 36&  .240 &  7& -.138 & 15& -.045 & 36&  .257 &  6& -.026 & 30& -.037 & 38&  .210 &  8& -.097 & 13\\ 
FreeThrowsMade         & -.187 & 15&  .022 & 35&  .219 &  6& -.192 & 12& -.002 & 40& -.034 & 25& -.188 & 11&  .001 & 39& -.031 & 27\\ 
FreeThrowsAttempted    & -.189 & 13& -.033 & 31&  .218 &  7& -.192 & 13& -.037 & 30& -.020 & 33& -.187 & 12& -.030 & 32& -.005 & 35\\ 
FreeThrowPercentage    & -.063 & 34&  .135 & 15& -.169 & 12& -.048 & 35&  .200 &  9& -.114 &  8& -.047 & 37&  .170 & 10& -.198 &  6\\ 
OffensiveRebound       & -.134 & 27& -.315 &  3& -.060 & 30& -.161 & 24& -.246 &  7&  .089 & 10& -.158 & 26& -.221 &  7&  .125 &  9\\ 
DefensiveRebound       & -.185 & 17& -.182 & 12& -.013 & 39& -.188 & 15& -.131 & 15&  .027 & 28& -.183 & 16& -.115 & 16&  .059 & 17\\ 
TotalRebound           & -.176 & 20& -.228 &  8& -.027 & 36& -.184 & 18& -.168 & 11&  .046 & 17& -.179 & 19& -.149 & 12&  .080 & 15\\ 
Assist                 & -.157 & 23&  .172 & 14&  .221 &  5& -.175 & 23&  .143 & 13& -.015 & 34& -.169 & 23&  .145 & 13& -.004 & 37\\ 
Turnover               & -.196 & 10&  .064 & 22&  .151 & 13& -.193 & 11&  .045 & 26& -.014 & 35& -.187 & 14&  .052 & 25&  .012 & 33\\ 
Steal                  & -.183 & 18&  .106 & 17&  .021 & 38& -.184 & 17&  .086 & 17&  .038 & 19& -.177 & 21&  .095 & 19&  .051 & 20\\ 
Block                  & -.122 & 30& -.281 &  4& -.027 & 37& -.152 & 25& -.231 &  8&  .052 & 16& -.149 & 27& -.210 &  9&  .091 & 14\\ 
PersonalFouls          & -.188 & 14& -.070 & 21& -.204 &  8& -.184 & 20& -.071 & 20&  .123 &  7& -.178 & 20& -.050 & 27&  .127 &  8\\ 
DoubleDouble           & -.137 & 26& -.207 & 10&  .287 &  4& -.143 & 26& -.185 & 10& -.094 &  9& -.164 & 25& -.157 & 11&  .002 & 40\\ 
TripleDouble           & -.067 & 33&  .012 & 36&  .404 &  1& -.071 & 33&  .008 & 38& -.181 &  5& -.198 &  1&  .001 & 40&  .030 & 29\\ 
Points                 & -.207 &  4&  .055 & 24&  .076 & 26& -.199 &  1&  .041 & 28& -.033 & 26& -.194 &  2&  .041 & 30& -.036 & 26\\ 
PointsTurnOver         & -.198 &  9&  .099 & 18&  .061 & 28& -.194 &  8&  .076 & 19& -.009 & 38& -.190 &  8&  .078 & 21& -.016 & 32\\ 
PointsOffRebound       & -.179 & 19& -.189 & 11& -.010 & 40& -.184 & 19& -.140 & 14&  .020 & 32& -.182 & 17& -.129 & 15&  .038 & 23\\ 
FastBreakPoints        & -.169 & 22&  .178 & 13&  .148 & 14& -.178 & 21&  .147 & 12&  .009 & 37& -.172 & 22&  .145 & 14& -.003 & 38\\ 
PointsInPaint          & -.191 & 12& -.128 & 16&  .111 & 22& -.190 & 14& -.109 & 16&  .011 & 36& -.185 & 15& -.095 & 20&  .045 & 21\\ 
OpponentsTOPoints      & -.205 &  6&  .035 & 30& -.129 & 16& -.197 &  6&  .037 & 31&  .055 & 13& -.190 &  7&  .048 & 29&  .057 & 18\\ 
Opponent2ndPts         & -.205 &  5&  .046 & 25& -.116 & 20& -.196 &  7&  .047 & 24&  .053 & 14& -.189 & 10&  .057 & 23&  .053 & 19\\ 
OpponentFastBreakPts   & -.202 &  8&  .040 & 28& -.126 & 17& -.194 & 10&  .039 & 29&  .053 & 15& -.187 & 13&  .049 & 28&  .059 & 16\\ 
OpponentPtsDuringPlay  & -.207 &  2&  .041 & 27& -.123 & 19& -.198 &  4&  .045 & 25&  .041 & 18& -.192 &  6&  .053 & 24&  .038 & 24\\ 
PlayerBlocked          & -.185 & 16& -.009 & 39&  .074 & 27& -.185 & 16& -.016 & 34&  .026 & 29& -.182 & 18& -.005 & 37&  .037 & 25\\ 
TimesFouled            & -.195 & 11& -.030 & 32&  .188 &  9& -.194 &  9& -.029 & 32& -.006 & 39& -.190 &  9& -.020 & 36&  .007 & 34\\ 
Height                 &  .002 & 40& -.345 &  2& -.113 & 21& -.002 & 40& -.384 &  1& -.036 & 24& -.006 & 40& -.377 &  1&  .004 & 36\\ 
Weight                 & -.004 & 39& -.351 &  1& -.046 & 33& -.005 & 39& -.374 &  2& -.037 & 21& -.008 & 39& -.369 &  2& -.002 & 39\\ 
draftRound             &  .078 & 32&  .045 & 26&  .085 & 25&  .077 & 32&  .049 & 23&  .597 &  2&  .104 & 32&  .101 & 18&  .543 &  1\\ 
draftNumber            &  .091 & 31&  .056 & 23&  .061 & 29&  .090 & 31&  .068 & 21&  .608 &  1&  .090 & 33&  .114 & 17&  .475 &  2\\ 
MVP                    & -.043 & 37&  .028 & 33&  .180 & 11& -.034 & 37&  .012 & 35& -.270 &  3& -.089 & 34& -.023 & 34& -.449 &  3\\ 
DefensiveMVP           & -.017 & 38& -.077 & 19&  .056 & 31& -.015 & 38& -.078 & 18& -.086 & 11& -.054 & 36& -.331 &  3& -.247 &  4\\ 

    \hline
  \end{tabular}
  \caption{Complete factors for rank-three decompositions of the
    basketball data set. The component rankings (in absolute value)
    are shown in parenthesis.}
  \label{tab:bbfacts}
\end{table}

\end{document}